\pdfoutput=1

\documentclass[11pt]{article}
\usepackage[]{acl}

\usepackage{times}
\usepackage{latexsym}
\usepackage{dblfloatfix}
\usepackage{multirow}
\usepackage{amsmath}
\usepackage[most]{tcolorbox}

\newcommand{\examplepOne}{1}
\newcommand{\examplepTwo}{2}
\newcommand{\examplepThree}{3}
\newcommand{\examplepFour}{4}
\newcommand{\examplepFive}{5}
\newcommand{\examplepSix}{6}
\newcommand{\examplepSeven}{7}

\usepackage{todonotes}
\definecolor{TodoColor}{rgb}{1,0.7,0.6}
\definecolor{TodoColor2}{HTML}{AACCAA}

\makeatletter\def\Hy@Warning#1{}\makeatother
\let\svthefootnote\thefootnote
\newcommand\blankfootnote[1]{%
  \let\thefootnote\relax\footnotetext{#1}%
  \let\thefootnote\svthefootnote%
}

\usepackage[T1]{fontenc}

\usepackage{parskip} 

\usepackage[utf8]{inputenc}
\usepackage[linesnumbered,ruled,vlined]{algorithm2e} 
\usepackage{microtype}

\usepackage{inconsolata}
\usepackage{graphicx}

\usepackage{booktabs}

\title{Autoformalizing Natural Language to First-Order Logic: \\A Case Study in Logical Fallacy Detection}

\author{
Abhinav Lalwani\textsuperscript{\rm 1}\thanks{Equal contribution} { }
Tasha Kim\textsuperscript{\rm 1}\footnotemark[1] { }
Lovish Chopra\textsuperscript{\rm 1}\footnotemark[1] { }
Christopher Hahn\textsuperscript{\rm 2}
\thanks{Work done while at Stanford University} { }
\\  
{\bf
Zhijing Jin\textsuperscript{\rm 3,4,5,\thanks{Co-supervision}} { } 
Mrinmaya Sachan\textsuperscript{\rm 4,\footnotemark[3]} { } 
}
\\
\textsuperscript{\rm 1}Stanford University { }
\textsuperscript{\rm 2}X, the moonshot factory {}
\\
\textsuperscript{\rm 3}Max Planck Institute for Intelligent Systems { } 
\textsuperscript{\rm 4}ETH Zürich { }
\textsuperscript{\rm 5}University of Toronto \\
\texttt{\{lalwani,tashakim,lovish\}@stanford.edu { }
\{jinzhi,msachan\}@ethz.ch}
}
\begin{document}
\maketitle
\begin{abstract}
Translating natural language into formal language such as First-Order Logic (FOL) is a foundational challenge in 
NLP with wide-ranging applications in automated reasoning, misinformation tracking, and knowledge validation. In this paper, we introduce Natural Language to First-Order Logic (NL2FOL), a framework to autoformalize natural language to FOL step-by-step using Large Language Models (LLMs). Our approach addresses key challenges in this translation process, including the integration of implicit background knowledge. By leveraging structured representations generated by NL2FOL, we use Satisfiability Modulo Theory (SMT) solvers to reason about the logical validity of natural language statements. We present logical fallacy detection as a case study to evaluate the efficacy of NL2FOL.  Being neurosymbolic, our approach also provides interpretable insights into the reasoning process and demonstrates robustness without requiring model fine-tuning or labeled training data. Our framework achieves strong performance on multiple datasets -- on the \textsc{Logic} dataset, NL2FOL achieves an F1-score of 78\%, while generalizing effectively to the \textsc{LogicClimate} dataset with an F1-score of 80\%.\footnote{Code available at: \url{github.com/lovishchopra/NL2FOL}}

\end{abstract}

\begin{table*}[ht]
\centering\small
\renewcommand{\arraystretch}{1.4}
\begin{tabular}{ 
p{0.17\linewidth} p{0.38\linewidth} p{0.34\linewidth}}

\toprule
\textbf{Fallacy Name} & \textbf{Example} & \textbf{Logical Form} \\
\midrule
Faulty Generalization &  Sometimes flu vaccines don’t work; therefore vaccines are useless. & $(\exists x \in \mathrm{FluVaccines} (\mathrm{DoesntWork}(x)) \wedge (\mathrm{FluVaccines} \subseteq \mathrm{Vaccines})) \Rightarrow $\newline$(\forall y \in \mathrm{Vaccines} \ (\mathrm{DoesntWork}(y)))$  \\ \hline
False Causality & Every time I wash my car, it rains. Me washing my car has a definite effect on the weather. & $ \mathrm{occuredAfter}(\mathrm{washingCar},\mathrm{rain}) \Rightarrow \mathrm{caused}(\mathrm{washingCar},\mathrm{rain})$
 \\ \hline
Ad Populum & Everyone should like coffee: 95\% of teachers do! & $(\mathrm{like}(\mathrm{coffee}, \mathrm{95\%Teachers})) 
\;\Rightarrow\;
(\mathrm{like}(\mathrm{coffee}, \mathrm{everyone}))$
 \\ \hline 
False Dilemma & I don’t want to give up my car, so I don’t think I can support fighting climate change. & $\forall(a) (\mathrm{giveUpCar}(a) \lor \mathrm{dontSupportFightingClimateChange}(a)) $ \\ \bottomrule
\end{tabular}
\caption{Sample logical fallacies from \citet{jin2022logical} along with examples and their logical forms. For each type of fallacy, we show one possible logical form.
}
\label{tab:each_fallacy_desc}
\vspace{-1.5em}
\end{table*}
\section{Introduction}

In recent years, Large Language Models (LLMs) have shown impressive advancements in understanding and generating natural language \cite{brown2020language}. Despite this progress, their ability to tackle complex reasoning tasks remains limited \cite{bubeck2023sparks,wei2022chain}. These challenges are especially prevalent in multistep logical deductions, abstract reasoning, and knowledge integration in various domains \cite{dalvi2021explaining,chen2024selfplay}. Addressing these limitations and improving the reasoning capabilities of LLMs has become a critical focus in AI research \cite{jin2023iterative,gendron2024largelanguagemodelsstrong}. 

In contrast, formal reasoning tools such as Satisfiability Modulo Theory (SMT) solvers excel in reasoning, providing rigorous, provable guarantees by leveraging symbolic representations and logical calculus \cite{barrett2009smt,demoura2008z3}. However, a key limitation of formal solvers is their reliance on structured logical input, such as First Order Logic (FOL), which must accurately capture the semantics and context of natural language statements \cite{beltagy2016representing}. This presents the challenge of translating unstructured natural language into a structured form required for formal reasoning while preserving essential context and meaning.

This also brings a unique opportunity: if we can reliably translate natural language into structured logical forms, we can harness the power of formal solvers to reason systematically over natural language statements. However, achieving this translation is nontrivial, as it involves accurately capturing natural language semantics \cite{beltagy2016representing}. Moreover, translating to a formal logical form may cause implicit and external context to be lost, which must be reintroduced to ensure logical accuracy.

To address these challenges, we present NL2FOL, a novel framework that bridges the gap between natural language and formal reasoning systems. NL2FOL employs a structured, step-by-step pipeline to translate natural language inputs into first-order logic (FOL) representations, leveraging large language models (LLMs) at each step for enhanced precision and adaptability. A distinguishing feature of NL2FOL is its seamless integration of background knowledge into the generated logical forms, overcoming a major limitation of traditional formal logic frameworks - the inability to capture implicit information embedded in natural language.

In this paper, we demonstrate the effectiveness of NL2FOL through a case study on logical fallacy detection, showcasing its ability to identify and explain faulty reasoning in natural language arguments. Detecting logical fallacies is particularly challenging as they often rely on reasoning patterns that appear plausible yet are fundamentally flawed \cite{jin2022logical}. To address this, NL2FOL translates logical fallacies from natural language into FOL representations, enabling formal solvers to verify logical validity. These solvers generate counterexamples and explanations, which are interpreted back into natural language to enhance human comprehensibility. By incorporating intermediate natural language outputs, our pipeline improves interpretability, transparency, and debuggability \cite{lin2021attention}.

We show that our framework achieves strong performance on the logical fallacy detection benchmarks \textsc{Logic} and \textsc{LogicClimate} \cite{jin2022logical}, with F1 scores of 78\% and 80\%, respectively - outperforming existing models by 22\% on the challenge set, \textsc{LogicClimate}. These results highlight NL2FOL as a generalizable and interpretable tool for reasoning tasks that demand the precision of formal reasoning systems. By analyzing the strengths and weaknesses of LLMs at each step of the NL2FOL pipeline, we further identify opportunities for improving logical reasoning capabilities. Even though LLMs prove to be effective in parsing and generating logical representations for structured inputs, they often struggle with ambiguities in natural language and incorporating nuanced contextual knowledge. The ability to integrate symbolic solvers with language models positions NL2FOL as a powerful neurosymbolic approach, bridging the gap between formal reasoning and natural language understanding.

\begin{figure*}[t]
\centering
\includegraphics[width=.90\linewidth]{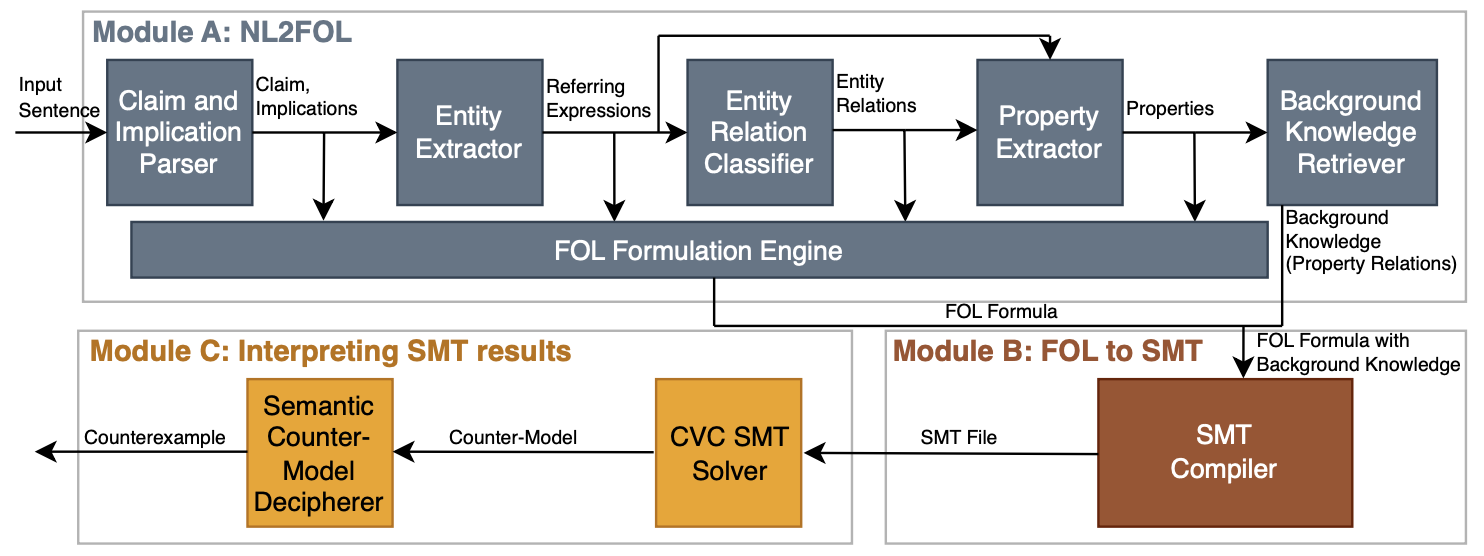}
\caption{Overview of the proposed framework used for logical fallacy detection. \textit{Module A} converts natural language input to a first-order logic formula merged with contextual relationships, \textit{Module B} compiles the negation of a given logical formula to an SMT file with well-defined sorts for variables and predicates, and \textit{Module C} runs CVC on the SMT file and if the negation is satisfiable, interprets the counter-model in natural language.
}
\label{fig:model}
\vspace{-1em}
\end{figure*}


 
\section{Related Work}
\textbf{Logical fallacy detection.} Existing work on classifying logical fallacies includes argument sufficiency classification \cite{b2}, ad hominem fallacies from Reddit posts \cite{b3} and dialogues \cite{b4}, rule parsers \cite{b5}, structure-aware Transformers \cite{jin2022logical}, multitask instruction based prompting \cite{alhindi-etal-2022-multitask}, and instance-based reasoning \cite{b12}. To our knowledge, our work is the first on few-shot classification of logical fallacies in a step-by-step, explainable manner. By ensuring that the reasoning process is transparent, we allow users to understand and verify the system decision.

\textbf{Natural language to formal logic.} While early work on mapping text to formal logic relied heavily on grammar-based approaches \cite{b29,b31,b32}, recent advances in deep learning and foundation models have enabled new data-driven techniques for translating natural language to linear temporal logic \cite{b10,b13,b14} and first-order logic \cite{b34,b22,hahn2022formal}. Neural models for parsing natural language to first-order logic \cite{b34,b22} and neuro-symbolic approach combining language models with first-order logic provers \cite{olausson-etal-2023-linc} have since been explored. However, these approaches still face challenges in accurately capturing implicit information or transforming complex ambiguous sentences into logical form, mainly attributed to linguistic ambiguity.

\citet{aly2023qanatver} integrated LLMs with logical inference for fact verification, and while our method shares the fundamental idea of employing LLMs to construct proofs and analyze relationships between textual spans, our task adds a layer of contextual reasoning by requiring the incorporation of background knowledge and maintaining interdependency between proof steps, which is not present in approaches where each proof step is treated as an independent, isolated process.

\textbf{Theory solvers.} Recent work by \citet{hahn2022formal} demonstrated the potential of integrating symbolic solvers with large language models (LLMs), such as tool-augmented LLMs, to combine neural and symbolic reasoning. While such approaches are promising, they often struggle to translate natural language into symbolic representations and effectively capture background knowledge.  Other recent approaches \citep{olausson-etal-2023-linc, pan2023logiclmempoweringlargelanguage} have used theory solvers to logically reason with natural language, which we build on with several key advancements. First, we introduce a framework that handles naturalistic, real-world data and tasks with ambiguous premises and conclusions. Then, we present a method to incorporate background knowledge into logical formulas. Finally, we show that our approach introduces interpretability by allowing human verification and modification throughout the intermediate reasoning steps.

\section{Methodology}
Although powerful, LLMs struggle to detect logical fallacies 
in language,
as it requires proper logical analysis \citep{jin2022logical}.
On the other hand, SMT solvers can reason over logical formulas with theoretical guarantees but require the input to be in a structured, logical form. This approach combines the strengths of both to classify logical fallacies.

{\bf Task formulation.} The task input is an argument in natural language comprising one or more sentences, which is converted into formal logical form using a chain of LLMs. Following this, an SMT solver processes the logical form and returns whether it is valid. If invalid, the SMT solver provides a counterexample explaining why it is a logical fallacy, which is then interpreted with an LLM.

{\bf First-order logic.} In FOL, propositions are represented using predicates that express properties or relations over objects in a domain. These predicates can be combined with constants, representing specific objects and variables that represent unspecified elements in the domain. An $\mathrm{Interpretation}$ assigns meaning to these symbols within a given context, while a $\mathrm{Sort}$ categorizes objects into different types, facilitating precise reasoning about their properties. Logical connectives of FOL, such as implication ($\Rightarrow$), universal quantifiers ($\forall$), existential quantifiers ($\exists$), and operators for conjunction/and ($\land$), disjunction/or ($\lor$), and negation/not ($\neg$), allow for the construction of intricate statements.


{\bf Module A: Natural language to first-order logic.}
Our approach for converting given natural language sentences into a logical form comprises multiple steps involving few-shot prompting of LLMs: (i) decomposing a sentence into multiple smaller parts that can be represented in first-order logic, (ii) identifying relationships between different sub-components to merge them and obtain a resultant logical formula, and (iii) identifying real-world relationships between these sub-components (background knowledge) and augmenting them to obtain a FOL formula by incorporating background knowledge in the statement. We demonstrate with a Logical Fallacy (LF) and a Valid (V) example.

\fbox{
\begin{minipage}{\dimexpr\columnwidth-3\fboxsep-2\fboxrule\relax}
\fontfamily{courier}\small
{\bf 1. LF Example: A Logical Fallacy Input}\\
I met a tall man who loved to eat cheese, now I believe all tall people like cheese. 
\vspace{5pt}
\\
{\bf 2. V Example: A Valid Input}\\
A boy is jumping on a skateboard in the middle of a red bridge. Thus the boy does a skateboarding trick.
\end{minipage}
}
\vspace{-5pt}

Our pipeline begins with a semantic decomposition module which decomposes natural language arguments into respective claims and implications. Generally, a sentence can be split into some claims and implications based on those claims (see Prompt~\hyperlink{ex:p2}{\examplepTwo}).


\fbox{
    \begin{minipage}{\dimexpr\columnwidth-4\fboxsep-2\fboxrule\relax}
    \fontfamily{courier}\small{
    \bf 1. LF Example: Claim and Implication Parser}\\
    \textit{Claim:} A tall man loved to eat cheese.\\
    \textit{Implication}: All tall people like cheese.
    \vspace{5pt}
    \\
    \fontfamily{courier}\small{\bf 2. V Example: Claim and Implication Parser}\\
    \textit{Claim:} A boy is jumping on a skateboard in the middle of a red bridge.\\
    \textit{Implication:} The boy does a skateboarding trick.
    \end{minipage}
}
\vspace{-5pt}

The claims and implications are split into further sub-components and used to 
build up the logical form of the sentence. The next step is to identify entities in the sentence. In our work, we treat 
noun phrases or surrogates for noun phrases as entities (see Prompt~\hyperlink{ex:p3}{\examplepThree}).
Then, we find the relationship between the different entities using Zero-Shot classification via Natural Language Inference (NLI). These relationships (e.g., subset, equality, not related) are generally helpful in deciding appropriate quantifiers in the logical form. For example, if the entities are \emph{man} and \emph{people}, then it can be inferred that \emph{man} is a subset of \emph{people} and that the man would be bound by an existential quantifier in the sentence \emph{x} (see Prompt~\hyperlink{ex:p4}{\examplepFour}).

\noindent
\fbox{
    \begin{minipage}{\dimexpr\columnwidth-4\fboxsep-2\fboxrule\relax}
    \fontfamily{courier}\small
    {\bf 1. LF Example: Entity Extractor}\\
    \textit{Referring expressions:} 
    \begin{itemize}
        \item man: x
        \item cheese: c
        \item people: y
        \item $x \subseteq y$\\
    \end{itemize}
\vspace{-5pt}
    {\bf 2. V Example: Entity Extractor}\\
    \textit{Referring expressions:}
    \begin{itemize}
        \item boy: b
        \item skateboard: s
        \item bridge
        \item skateboardingTrick: y
    \end{itemize}
    \end{minipage}
}
\vspace{-5pt}

The other set of sub-components are properties, which describe a trait of a referring expression or relationship between multiple referring expressions. 
These properties are predicates in first-order logic. We use a single module to extract the properties and the relation between properties and entities.  (see Prompt~\hyperlink{ex:p5}{\examplepFive}). We also find the relationships between various properties  (see Prompt~\hyperlink{ex:p6}{\examplepSix}). For instance, in the LF Example, it can be inferred that \emph{Like} and \emph{Love} are contextually similar. Similarly, in our valid example, \emph{jumping over skateboard} implies \emph{doing a skateboard trick}. These relationships provide an additional 
context that is not directly present in the statement.

To identify these contextual relationships, we run NLI between each pair of properties, i.e., by setting one property as the hypothesis and the other as the premise as the input to the NLI model. If we find that any one property entails the other, we add the relationship $\mathrm{property1} \Rightarrow \mathrm{property2}$ to our context. Before running the NLI model between a pair of properties, we replace the variables in each property with the referring expressions that they represent. This adds additional context that helps the NLI model identify relations. For instance, in the V Example, the NLI model is unable to find the relation between \textit{JumpsOn(x, s)} and \textit{Does(x, y)}, but it can identify the relationship between \textit{JumpsOn(boy, skateboard)} and \textit{Does(boy, skateboardingTrick)}.

\fbox{
\begin{minipage}{\dimexpr\columnwidth-4\fboxsep-2\fboxrule\relax}
\fontfamily{courier}\small
{\bf 1. LF Example: Property Extractor + Background Knowledge Retriever}\\
\textit{Properties:} Tall, Love, Like \\
\textit{Property entity relations:} $\mathrm{Tall}(x), \mathrm{Love}(x, c)$ \\
\textit{Background knowledge:}
\begin{enumerate}
    \item $\forall x (\mathrm{Like}(x, c) \Rightarrow \mathrm{Love}(x, c))$
    \item $\forall x (\mathrm{Love}(x, c) \Rightarrow \mathrm{Like}(x, c))$
    \item $x \subseteq y$\\
\end{enumerate}
\vspace{-5pt}
{\bf 2. V Example: Property Extractor + Background Knowledge Retriever}\\
\textit{Properties:} JumpsOn, inMiddleOf, Red, Does \\
\textit{Property entity relations:} $\mathrm{JumpsOn}(b, s),$\\
$\mathrm{Red}(\mathrm{bridge}),
\mathrm{inMiddleOf}(b, \mathrm{bridge}), \mathrm{Does}(b, y)$ \\
\textit{Background knowledge:}
\begin{enumerate}
    \item $\forall x (\mathrm{JumpsOn}(b, s) \Rightarrow \mathrm{Does}(b, y))$
\end{enumerate}
\end{minipage}
}
\vspace{-5pt}

Finally, we combine all of this information using the relationships between properties and entities to obtain the FOL form of the sentence with the help of an LLM (see Prompt~\hyperlink{ex:p7}{\examplepSeven}). For a logical fallacy, the negation of the formula is expected to be satisfiable. On the contrary, for a valid statement, the negation of the formula should be unsatisfiable.

\fbox{
\begin{minipage}
{\dimexpr\columnwidth-4\fboxsep-2\fboxrule\relax}
\fontfamily{courier}\small
{\bf 1. LF Example: NL2FOL Output}\\
\textit{First-order logic:}
$((\forall x (\mathrm{Like}(x,c) \Rightarrow \mathrm{Love}(x,c))) \wedge $ 
$(\forall x (\mathrm{Love}(x,c) \Rightarrow \mathrm{Like}(x,c))) \wedge $ 
$ (\exists x (\mathrm{Tall}(x) \wedge \mathrm{Love}(x,c)))) \Rightarrow$ 
$ (\forall y (\mathrm{Tall}(y) \Rightarrow \mathrm{Like}(y,c)))$ 
\vspace{5pt}
\\
\fontfamily{courier}\small
{\bf 2. V Example: NL2FOL Output}\\
\textit{First-order logic:}
$(\forall x (\mathrm{JumpsOn}(x,s) \Rightarrow \mathrm{Does}(x,y)) \wedge \mathrm{Red}(\mathrm{bridge}) \wedge$ 
$ \mathrm{inMiddleOf}(b, \mathrm{bridge}) \wedge \mathrm{JumpsOn}(b,s)) \Rightarrow$ 
$\mathrm{Does}(b,y)$
\end{minipage}
}

{\bf Module B: First-order logic to SMT.}
The next step involves automatically creating an SMT file for the negation of the first-order logical formula generated. While one can easily write an SMT file for a logical formula manually, generating one automatically for an arbitrary formula has not been done before.  Thus, we develop a compiler that parses a given logical formula and converts it into an SMT file that can be given to CVC as input, as described in Algorithm \ref{alg:foltosmt} (See Appendix). 

\begin{table*}[t]
\centering
\small
\renewcommand{\arraystretch}{1} 
\begin{tabular}{llcccccccc}
\toprule[1pt]
 & 
& \multicolumn{4}{c}{\textsc{Logic}} 
& \multicolumn{4}{c}{\textsc{LogicClimate}} \\
\cmidrule(lr){3-6} \cmidrule(lr){7-10}
\textbf{Model}& \textbf{Method} & \textbf{Acc.} & \textbf{P.} & \textbf{R.} & \textbf{F1} 
  & \textbf{Acc.} & \textbf{P.} & \textbf{R.} & \textbf{F1} \\
\toprule[1pt]
\multirow{2}{*}{Llama-7B} 
 & End-to-end     & 0.41 & 0.45 & 0.82 & 0.58 & 0.31 & 0.38 & 0.62 & 0.47 \\
 & NL2FOL (Ours)  & {\bf 0.63} & {\bf 0.58} & {\bf 0.92} & {\bf 0.71} & {\bf 0.66} & {\bf 0.60} & \textbf{0.94} & \textbf{0.73} \\
\midrule
\multirow{2}{*}{GPT-4o-mini} 
 & End-to-end     & {\bf 0.91} & {\bf 0.94} & 0.88 & {\bf 0.91} & 0.64 & {\bf 0.67} & 0.55 & 0.60 \\
 & NL2FOL (Ours)  & 0.70 & 0.64 & {\bf 0.91} & 0.75 & {\bf 0.73} & 0.66 & {\bf 0.93} & {\bf 0.77} \\
\midrule
\multirow{2}{*}{GPT-4o} 
 & End-to-end     & \textbf{0.96} & \textbf{0.96} & {\bf 0.96} & {\bf 0.96} 
                  & 0.70 & \textbf{0.95} & 0.42 & 0.58 \\
 & NL2FOL (Ours)  & 0.78 & 0.76 & 0.82 & 0.78 
                  & \textbf{0.80} & 0.80 & {\bf 0.80} & {\bf 0.80} \\
\midrule
\multirow{2}{*}{OpenAI o1-preview} 
 & End-to-end     & 0.93 & 0.89 & 0.98 & 0.93 
                  & 0.73 & 0.84 & 0.56 & 0.67 \\
 & NL2FOL (Ours)  & -    & -    & -    & -    & -    & -    & -    & - \\
\bottomrule[1pt]
\end{tabular}
\caption{Comparison of few-shot model performance metrics (abbreviations: Acc. = accuracy, P. = precision, R. = recall, F1 = F1 score) on the \textsc{Logic}+SNLI and \textsc{LogicClimate}+SNLI datasets using End-to-end vs.\ NL2FOL (Ours). Results on NL2FOL with o1-preview are omitted as o1-preview failed to complete the pipeline in most cases, likely due to its poor instruction following capabilities.}
\label{tab:combined-logic}
\vspace{-1.5em}
\end{table*}

{\bf Module C: Interpreting SMT results.} To verify the validity of the logical formulas, we utilize an SMT solver, CVC4 \cite{b25}. The solver determines whether the formula is valid or invalid, hence a logical fallacy. In the case of invalidity, the model provides a counterexample to the original logical formula, which shows that the given claim or implication is a logical fallacy.
\vspace{-0.5em}
\begin{figure}[h]
    \centering
    \fbox{
        \begin{minipage}{\dimexpr\columnwidth-4\fboxsep-2\fboxrule\relax}
        \fontfamily{courier}\small
        {\bf Example (Module B Output):}\\
        I met a tall man who loved to eat cheese, now I believe all tall people like cheese.\vspace{-1em}
        \[
        \vspace{-1em}\downarrow
        \]
        \textit{First-order logic:}
        $((\forall x (\mathrm{Like}(x, c) \Rightarrow \mathrm{Love}(x, c))) \wedge $ 
        $(\forall x (\mathrm{Love}(x, c) \Rightarrow \mathrm{Like}(x, c))) \wedge $ 
        $ (\exists x (\mathrm{Tall}(x) \wedge \mathrm{Love}(x, c)))) \Rightarrow$ 
        $ (\forall y (\mathrm{Tall}(y) \Rightarrow \mathrm{Like}(y, c)))$
        \vspace{-1em}\[
        \vspace{-1em}\downarrow
        \]
        SMT classification: Logical fallacy\\
        Explanation: Counterexample\vspace{-1em}
        \[
        \vspace{-1em}\downarrow
        \]
        \begin{itemize}
            \item John is tall ($\mathrm{Tall}(\mathrm{John})$ is True). John likes cheese ($\mathrm{Likes}(\mathrm{John}, \mathrm{Cheese}$) is True).
            \item Jane is tall ($\mathrm{Tall}(\mathrm{Jane})$ is True). No constraint Jane likes cheese.\\
        \end{itemize}
        Therefore, there exists a tall person (John) who likes cheese, but it does not follow that all tall people like cheese, since Jane serves as a counterexample.
        \end{minipage}
        }
    \caption{Example of logical fallacy detection using NL2FOL. The resulting classification is explained using a counterexample generated by the SMT solver.}
    \label{fig:counterexample}
\vspace{-1em}
\end{figure}

The result of the SMT solver is hard to interpret, as it uses technical terminology 
generally only well understood by those who are familiar with $\mathrm{CVC4}$ and SMT. To obtain an explanation in natural language, we prompt an LLM with the claim, implication, referring expressions, properties, FOL formula, and the counterexample generated by $\mathrm{CVC4}$. The model then interprets the counterexample with natural language, as depicted in Figure \ref{fig:counterexample}.

\section{Experiments}

We evaluate our approach on both logical fallacies (positive class) and valid statements (negative class). For logical fallacies, we use the \textsc{Logic} and \textsc{LogicClimate} \cite{jin2022logical} datasets, originally designed for training models to identify and classify different fallacies. These datasets contain examples of logical fallacies, each labeled with multiple categories from 13 different categories, including faulty generalization, circular claim, and ad hominem. The  \textsc{Logic} dataset contains 2,449 examples of common logical fallacies collected mostly from quiz websites. The \textsc{LogicClimate} dataset comprises 1,079 examples of logical fallacies drawn from climate change news articles on the Climate Feedback platform. It is intended to test the model's ability to generalize out-of-domain.

To test our approach with valid statements, we use the Stanford Natural Language Inference (SNLI) corpus \cite{b21}, which supports the development of natural language inference systems. This dataset features over 570,000 human-annotated sentence pairs, where each pair consists of a premise and a hypothesis labeled as entailment, contradiction, or neutral. We focus on the entailment class in this study, extracting over 170,000 sentence pairs where the premise entails the hypothesis. We construct valid sentences by combining the premise and hypothesis into a single sentence.

The task is set up as a simple binary classification task, where the input consists of sentences drawn from the \textsc{Logic} or \textsc{LogicClimate} datasets labeled as logical fallacies or from the SNLI dataset labeled as valid sentences. Here, we treat logical fallacies as the positive class. To ensure a balanced evaluation, we select an equal number of fallacies and valid statements, allowing for a fair comparison across both classes.
Finally, our model is evaluated on standard binary classification metrics such as precision, recall, f1 score, and accuracy.

\textbf{Models.} We compare our method to pretrained language models, including Llama2-7B \cite{b26}, GPT4o-mini \cite{4omini}, GPT4o \cite{openai2024gpt4ocard} and OpenAI o1-preview \cite{openai2024openaio1card} with few-shot in-context examples (see Prompt~\hyperlink{ex:p1}{\examplepOne}). We also run NL2FOL with each of the above models used for the LLM prompting stages. Llama2-7B was chosen for our experiments as it had the best performance during testing over an initial subset of the data, outperforming Llama3.1-8B \cite{grattafiori2024llama3herdmodels}, Llama3.2-11B \cite{llama11b}, and Ministral-8B \cite{mistral8b}. We evaluate BART (140M parameters) \cite{b20} finetuned on MNLI \cite{williams-etal-2018-broad} to analyze the relationships between properties and referring expressions. 
We ran the experiments on a V100 GPU, with one run costing around 2 GPU hours.

\textbf{Prompt tuning.} 
For prompt tuning, 20 samples from the \textsc{Logic} dataset were selected and manually annotated with intermediate and final results. They were then split into 10 train and 10 validation examples. For each prompt, we start with a simple description of the task. 4-6 examples were randomly selected from the train set as in-context examples, with the relevant intermediate outputs depending on the stage. Results were tested on the validation examples, and the prompt was updated to address common mistakes. To ensure fairness, a fixed number of 5 improvement iterations was used for each prompt, and the one showing best performance over the validation examples was chosen.


{
\renewcommand{\arraystretch}{1.3}
\begin{table*}[t]  
    \centering \small
    \begin{tabular}{p{0.04cm} p{0.4cm} p{2.5cm}p{6.1cm}p{4.4cm}}
        \toprule
        & \textbf{Type} & \textbf{Sentence} & \textbf{Logical Form} & \textbf{Prediction} \\
        \hline
        1 & LF & X has been around for years now. Y is new. Therefore, Y is better than X.	& (IsNew(a) $\wedge$ $\sim$ IsNew(b)) $\Rightarrow$ (IsBetterThan(a,b)) & LF: Correct prediction \\
        \hline
        2 & LF &
        Everyone is doing the Low-Carb Diet. &
        ($\exists$ b ($\exists$ a (IsDoing(b,a)))) $\Rightarrow$ ($\exists$ c ($\exists$ a (IsDoing(c,a)))).
        &
        V: Incorrect prediction - Wrong translation given when no claim given\\
        \hline
        3 & V &
        Two dogs are fighting in a field. Consequently, the two dogs are outside. &
        ($\exists$ b ($\exists$ a (IsFighting(a, b) $\wedge$ IsInField(b) $\wedge$ IsInField(b)))) $\Rightarrow$ ($\exists$ a (IsOutside(a)))
        &
        LF: Incorrect prediction - Missing semantic ground truth claim: $\forall$ a (IsInField(a) $\Rightarrow$ IsOutside(a))
        \\
        \hline
        4 & 
        V &
        A baseball player gets ready to catch a fly ball near the outfield fence. Therefore, a person is playing baseball outdoors.&
        ($\exists$ a (IsGettingReady(a) $\wedge$ (IsABaseballPlayer(a) $\wedge$ IsCatchingFlyBall(a) $\wedge$ IsNearOutfieldFence(a))) $\wedge$ ($\forall$ e ( IsABaseballPlayer(e) $\Rightarrow$ IsPlayingBaseball(e))) $\wedge$ ($\forall$ f ( IsPlayingBaseball(f) $\Rightarrow$ IsABaseballPlayer(f))) $\wedge$ ($\forall$ g ( IsNearOutfieldFence(g) $\Rightarrow$ IsOutdoors(g)))) $\Rightarrow$ ($\exists$ c ($\exists$ a (IsPlayingBaseball(a) $\wedge$ IsOutdoors(c)))) &
        V: Correct Prediction - The method identifies additional context by establishing relationships such as \textit{IsBaseballPlayer} implying \textit{IsPlayingBaseball}, and \textit{IsNearOutfieldFence} implying \textit{IsOutdoors}. 
        \\
        \hline
        5 &
        V &
        A woman sits alone on a park bench in the sun. Hence, a woman is in a park.	
            &
        (IsSittingOn(a, b) $\wedge$ isParkBench(b) $\wedge$ IsInSun(a)) $\Rightarrow$ (IsInPark(a)).	&
        LF: Incorrect prediction - Missing semantic ground truth claim: $\forall a \forall b$ (IsSittingOn(a, b) $\wedge$ isParkBench(b) $\Rightarrow$ IsInPark(a)) \\
        \hline
       6 & V & A woman is standing at a podium. Thus, a person is at a podium. &
    ($\exists a \exists b$ (IsStandingAt(b, a)) $\wedge$ $\forall f \forall e \forall d$ (IsStandingAt(d,e) $\Rightarrow$ IsAt(f,e)) $\Rightarrow \exists c \exists a$ (IsAt(c, a)) &
    V: Correct prediction - The method identifies additional context by establishing the relationship \textit{IsStandingAt} implying \textit{IsAt}.  \\
    \bottomrule
    \end{tabular}
    \caption{Some example outputs of our model (abbreviations: LF = Logical Fallacy, V = Valid statement)}\label{tab:examples}
\vspace{-1em}
\end{table*}

\section{Results and Discussion}
As shown in Table \ref{tab:combined-logic}, 
our method achieves an F1 score of 78\% when used with GPT-4o on the \textsc{Logic} dataset. When run end-to-end, the Llama-7B model reached an F1 score of only 58\%, but when used with the NL2FOL pipeline, reached a score of 71\%. Although end-to-end classification has shown better performance in other models, comparisons can be skewed because they may have been exposed to the \textsc{Logic} dataset and its labels during training because this dataset was compiled from publicly accessible web sources. On average, NL2FOL demonstrated high recall, whereas end-to-end classification demonstrated high precision. 

Our challenge set \textsc{LogicClimate}+SNLI contains real-world logical fallacies from climate change news. Since this dataset was used to test generalization, the in-context examples we provide to all models are from the \textsc{Logic} dataset. NL2FOL yields results that are highly similar to the results from \textsc{Logic}, whereas end-to-end classification saw a drop in performance. This demonstrates that our system is also robust and adapts well to real-world texts, including texts with significant domain-specific context. This makes it effective in detecting and mitigating misinformation. Specifically, on this dataset, we find that NL2FOL outperforms direct translation with all LLMs that we tested.
\vspace{-1.5px}

\subsection{Quantitative Analysis}

{\bf Error analysis and interpretability.}
The proposed method is interpretable due to the use of natural language inputs and outputs at each step of the pipeline. This structure allows for precise identification of the specific module responsible for a failure by examining intermediate results. To evaluate this aspect, we performed an in-depth error analysis by annotating the module responsible for failure in 100 incorrect predictions made by the model. The results 
are summarized in Table \ref{tab:mistakes}.

\begin{table}[h!]
    \centering
    \renewcommand{\arraystretch}{1}
    \resizebox{\columnwidth}{!}{\begin{tabular}{lc}
    \toprule
    \textbf{Sub-Module with Error} & \textbf{Error Proportion}\\
    \midrule
    Claim and Implication Parser         & 0.19\\
    Incorrect Label                      & 0.01\\
    Property Extractor                   & 0.13\\
    Background Knowledge Retriever          & 0.54\\
    FOL Formulation Engine & 0.13\\  
    \bottomrule
    \end{tabular}}
    \caption{Categorization of model errors by type on NL2FOL (GPT-4o), based on a review by domain experts in the logic of 100 randomly sampled examples}
\label{tab:mistakes}
\vspace{-1em}
\end{table}

Our analysis reveals that the majority of errors occur in the `Background Knowledge Retriever', involving missed or incorrectly added contextual information in the logical form. Other errors typically pertain to incorrect identification of claims, implications, or properties. In contrast, inaccuracies in the generation of logical forms are relatively infrequent, suggesting that the model performs well in constructing accurate logical representations when provided with reliable information about the constituent entities and properties within a sentence. This finding underscores the importance of improving the background knowledge retriever module to improve overall model performance.

\vspace{-0.2em}
{\bf Impact of adding background knowledge to NL2FOL.}
Based on the error analysis, missing or incorrect background knowledge was a significant contributor to incorrect predictions of our method. To quantitatively assess the impact of grounding on model performance, we evaluated several approaches for NLI in the Background Relation Extractor. These included: (a) a pipeline without any background knowledge as a baseline, (b) a model without context where the LLM (GPT4o) only processes the input properties, (c) an LLM that incorporates both the input sentence and properties and (d) a smaller model specifically fine-tuned for NLI (BART-MNLI). Results are presented in Table \ref{tab:compressed}.

We see that precision and recall both improve significantly with better grounding techniques. The LLM model with sentence context achieves the highest overall performance. This is likely due to the sentence context providing information about clauses that are omitted due to the choice of representation in FOL. This indicates that integrating robust grounding mechanisms is critical to enhancing the accuracy and reliability of the  method.
\vspace{-0.5em}

\begin{table}[h]
    \centering \small\setlength\tabcolsep{3pt}
    \renewcommand{\arraystretch}{1} 
    \resizebox{\columnwidth}{!}{
    \begin{tabular}{@{}lcccccccc@{}}
    \toprule
     & \multicolumn{4}{c}{\textsc{LOGIC+SNLI}} & \multicolumn{4}{c}{\textsc{LogicClimate+SNLI}} \\
    \cmidrule(lr){2-5} \cmidrule(lr){6-9}
     \textbf{Method} & \textbf{Acc.} & \textbf{P.} & \textbf{R.} & \textbf{F1} & \textbf{Acc.} & \textbf{P.} & \textbf{R.} & \textbf{F1} \\
    \midrule
    (a) No Grounding & 0.54 & 0.52 & \textbf{0.88} & 0.66 & 0.57 & 0.54 & \textbf{0.94} & 0.69 \\
    (b) LLM & 0.76 & \textbf{0.78} & 0.74 & 0.75 & 0.79 & 0.80 & 0.78 & 0.79 \\
    (c) LLM w/ context & \textbf{0.78} & 0.76 & 0.82 & \textbf{0.78} 
                  & \textbf{0.80} & 0.80 & 0.80 & \textbf{0.80} \\
    (d) BART-MNLI & 0.71 & 0.71 & 0.70 & 0.70 & 0.77 & \textbf{0.81} & 0.71 & 0.77 \\
    \bottomrule
    \end{tabular}}
    \caption{Comparison of different grounding methods on NL2FOL (GPT4o-mini) across the LOGIC+SNLI and LogicClimate+SNLI datasets}
    \label{tab:compressed}
    \vspace{-0.5em}
\end{table}

{\bf Impact of using an SMT solver.}
To assess the impact of using an SMT solver in our pipeline, we compared its performance against an LLM as a baseline for classifying the logical forms as valid or fallacies. The results, summarized in Table \ref{tab:combined}, demonstrate a significant improvement in performance metrics with the integration of the SMT solver. Results reveal the SMT-based approach significantly outperforms the LLM-based approach in all metrics across both the \textsc{Logic} and \textsc{LogicClimate} datasets. This underscores the advantage of formal reasoning systems like SMT solvers for tasks requiring precise logical inference and structured reasoning compared to LLMs, which may lack systematic consistency in such contexts.

\subsection{Qualitative Analysis}
\subsubsection{Success Modes of NL2FOL}
\vspace{2pt}
\textbf{S1: Captures implicit information not mentioned in premises.} Previous works that directly translate natural language to logical forms suffer from an inability to capture implicit information not mentioned in the premises \cite{olausson-etal-2023-linc}. Our ``Background Knowledge Retriever'' step allows us to capture this information in the final logical form. An  illustration of this can be found in Example 4 of Table \ref{tab:examples}.

\textbf{S2: Captures explicit information that is missed in the representation.}
Our pipeline is also able to capture information that is explicitly mentioned in the premises but missed due to the choice of representation in logical form. In Example 6, in Table \ref{tab:examples}, the fact that the woman is both standing and is at the podium is lost due to the choice representation \textit{IsStandingAt}. However, the fact that the woman is at the podium is recovered in the final logical form due to the identified background knowledge \textit{IsStandingAt} implies \textit{IsAt}.

\textbf{S3: Comparison to direct translation.} To evaluate the efficacy of the multi-step LLM pipeline, we compared it against a direct translation approach, where natural language inputs were converted into logical forms with a single LLM call using a few-shot prompt. However, this task proved to be excessively complex for LLMs. Llama failed to generate any output, citing an inability to comprehend the prompt. 
Larger LLMs exhibited significant limitations, with over 95\% of their outputs containing syntax errors. These findings highlight the inadequacy of direct translation for complex logical reasoning tasks and underscore the necessity of a structured, multi-step approach to ensure the accuracy and syntactic correctness of the logical form.

\subsubsection{Failure Modes of NL2FOL}
\vspace{2pt}
\textbf{F1: Misses some background knowledge.} As can be observed in Table \ref{tab:mistakes}, incorrect identification of background knowledge is the most common cause for incorrect classifications. This is because any gaps in background knowledge can cause a valid statement to be identified as a logical fallacy, and an incorrectly added clause can cause a fallacy to be identified as valid.
One such case is present in example 3 of the Table \ref{tab:examples}. In this case, the model is not able to identify the extra context statement because the NLI model does not identify a required ground-truth relation. If this context were to be added to the claim of the logical formula, then the statement would have been predicted to be valid.

\textbf{F2: Limitations of NLI.} Our current approach is limited to discerning relationships between two properties at a time rather than handling multiple relationships concurrently. For reference, consider Example 5 in Table \ref{tab:examples}. Here, the semantic claim involves the conjunction of two properties entailing the third, while the `Background Knowledge Retriever' only checks whether one property entails the other. Finding such complex extra context requires more advanced techniques or additional human intervention. Including them could further improve the precision of the model overall.

\textbf{F3: Imprecision of LLMs.}
Among the logical fallacies that our model incorrectly predicted to be a valid statement, most of these predictions failed due to the imprecision of the LLM, leading to false translations and incorrect results. Example 2 demonstrates a case where the input does not have any claim but instead jumps straight to an implication. However, the model is not able to identify that the example has no claim. As a result, we obtain an incorrect translation with our technique.

\begin{table}
\centering \small
\renewcommand{\arraystretch}{0.6}
    \setlength\tabcolsep{2pt}
\resizebox{\columnwidth}{!}{%
\begin{tabular}{@{}lcccccccc}
\toprule
 & \multicolumn{4}{c}{\textsc{Logic}} & \multicolumn{4}{c}{\textsc{LogicClimate}} \\
\cmidrule(lr){2-5} \cmidrule(lr){6-9}
 \textbf{Classifier} & \textbf{Acc.} & \textbf{P.} & \textbf{R.} & \textbf{F1} & \textbf{Acc.} & \textbf{P.} & \textbf{R.} & \textbf{F1} \\
\midrule
SMT &  \textbf{0.78} & \textbf{0.76} & \textbf{0.82} & \textbf{0.78} 
                  & \textbf{0.80} & \textbf{0.80} & \textbf{0.80} & \textbf{0.80} \\
GPT-4o & 0.69 & 0.71 & 0.62 & 0.66 & 0.73 & 0.72 & 0.74 & 0.73 \\
\bottomrule
\end{tabular}
}
\caption{Comparison of classification methods used with NL2FOL (GPT4o) on \textsc{Logic} and \textsc{LogicClimate}}\label{tab:combined}
\vspace{-1.6em}
\end{table}

\section{Conclusion}
We present an effective and automatic solution to detect fallacies and tackle
misinformation. We developed a strategy to distinguish logical fallacies from valid statements,
involving a chaining approach to convert a sentence to first-order logic using LLMs, followed by using SMT solvers to identify whether the first-order logical statement is valid or not. If not, we interpret the counter-model generated by the SMT solver in natural language. Our proposed technique shows promising results in identifying logical fallacies and valid statements, as well as good generalizability across domains.

\section*{Ethics Statement}

While the intended outcome of this research is to help fight misinformation and promote rational discourse, there are several ethical challenges that we must consider. First, dependence on AI to identify logical fallacies could influence how individuals engage in debates and discussions. There is a risk that people may over-rely on AI judgments, potentially stifling complex statements or dissenting opinions that are essential for a healthy democratic process. Moreover, the use of AI in moderating discussions, especially in identifying logical fallacies, raises ethical questions about the automation of content moderation. While it can enhance the quality of public discourse by filtering out fallacious statements, it also risks automating censorship and impacting the dynamics of online communities. In the wrong hands, logical fallacy detection tools could be exploited to silence speech or suppress viewpoints under the pretext of promoting rational discourse. This potentially allows governments or organizations to stifle opposition or critique.
 
To address these issues, we advocate for the development of ethical guidelines for AI use that emphasize transparency, accountability, and active user engagement. These measures are crucial in encouraging public literacy in AI and logical fallacies, ultimately empowering individuals to critically assess both AI output and arguments they may encounter. 

\section*{Limitations}
\textbf{Scope of logical reasoning tasks.} Correct identification of background knowledge is crucial for our method. While we have shown its potential in detecting logical fallacies for short and structured premises, it is important to note that this approach may miss complex relational constructs (for example, $(a \land b) \Rightarrow (c \vee d)$)), in which richer logical patterns may often be required in real-world reasoning tasks such as those present in multi-paragraph contexts or Question-Answering (QA) datasets.

\textbf{Generalizability to other tasks and domains.} We have demonstrated promising results of our approach to logical fallacy detection, but whether the findings generalize to other logical tasks and domains remains unexplored. The performance of our approach in other languages is untested and may introduce unforeseen challenges.

\textbf{Going beyond first-order logic.} It is unknown whether our approach would be sufficiently expressive for reasoning tasks requiring higher-order or non-classical logic, as we limit our exploration to first-order logic. Conceptually, extending our method to the aforementioned domains is feasible but would require modification to the SMT integration and LLM-driven logic translation processes. Thus, further testing may include translating to logic beyond FOL, such as temporal and higher-order logic.

\textbf{Computational cost.} Using LLMs and SMT solvers can incur high computational costs, such as high-performance GPUs for LLM inference, CPUs optimized for SMT solvers, and high API usage, particularly for models like GPT-o1 and Llama-7B.



\bibliography{ijcai24}

\begin{thebibliography}{41}
\expandafter\ifx\csname natexlab\endcsname\relax\def\natexlab#1{#1}\fi

\bibitem[{AI(2024{\natexlab{a}})}]{llama11b}
Meta AI. 2024{\natexlab{a}}.
\newblock \href {https://huggingface.co/meta-llama/Llama-3.2-11B-Vision} {Llama 3.2-11b model card}.
\newblock Accessed: 2025-02-15.

\bibitem[{AI(2024{\natexlab{b}})}]{mistral8b}
Mistral AI. 2024{\natexlab{b}}.
\newblock \href {https://huggingface.co/mistralai/Ministral-8B-Instruct-2410} {Ministral-8b-instruct-2410 model card}.
\newblock Accessed: 2025-02-15.

\bibitem[{Alhindi et~al.(2022)Alhindi, Chakrabarty, Musi, and Muresan}]{alhindi-etal-2022-multitask}
Tariq Alhindi, Tuhin Chakrabarty, Elena Musi, and Smaranda Muresan. 2022.
\newblock \href {https://doi.org/10.18653/v1/2022.emnlp-main.560} {Multitask instruction-based prompting for fallacy recognition}.
\newblock In \emph{Proceedings of the 2022 Conference on Empirical Methods in Natural Language Processing}, pages 8172--8187, Abu Dhabi, United Arab Emirates. Association for Computational Linguistics.

\bibitem[{Aly et~al.(2023)Aly, Strong, and Vlachos}]{aly2023qanatver}
Rami Aly, Marek Strong, and Andreas Vlachos. 2023.
\newblock \href {https://openreview.net/forum?id=UXSqUOMwbE} {{QA}-natver: Question answering for natural logic-based fact verification}.
\newblock In \emph{The 2023 Conference on Empirical Methods in Natural Language Processing}.

\bibitem[{Angeli and Manning(2014)}]{b31}
Gabor Angeli and Christopher~D Manning. 2014.
\newblock Naturalli: Natural logic inference for common sense reasoning.
\newblock In \emph{Proceedings of the 2014 Conference on Empirical Methods in Natural Language Processing (EMNLP)}, pages 534--545.

\bibitem[{Bai et~al.(2020)Bai, Liang, Zhang, Li, Bai, and Wang}]{lin2021attention}
Bing Bai, Jian Liang, Guanhua Zhang, Hao Li, Kun Bai, and Fei Wang. 2020.
\newblock \href {https://api.semanticscholar.org/CorpusID:235314008} {Why attentions may not be interpretable?}
\newblock \emph{Proceedings of the 27th ACM SIGKDD Conference on Knowledge Discovery \& Data Mining}.

\bibitem[{Barrett et~al.(2011)Barrett, Conway, Deters, Hadarean, Jovanovi{\'{c}}, King, Reynolds, and Tinelli}]{b25}
Clark Barrett, Christopher~L. Conway, Morgan Deters, Liana Hadarean, Dejan Jovanovi{\'{c}}, Tim King, Andrew Reynolds, and Cesare Tinelli. 2011.
\newblock Cvc4.
\newblock In \emph{Computer Aided Verification}, pages 171--177, Berlin, Heidelberg. Springer Berlin Heidelberg.

\bibitem[{Barrett et~al.(2009)Barrett, Stump, and Tinelli}]{barrett2009smt}
Clark Barrett, Aaron Stump, and Cesare Tinelli. 2009.
\newblock Satisfiability modulo theories.
\newblock \emph{Communications of the ACM}, 52(9):69--77.

\bibitem[{Beltagy et~al.(2016)Beltagy, Roller, Cheng, Erk, and J.~Mooney}]{beltagy2016representing}
Iz~Beltagy, Stephen Roller, Pengxiang Cheng, Katrin Erk, and Raymond J.~Mooney. 2016.
\newblock Representing meaning with a combination of logical and distributional models.
\newblock \emph{Computational Linguistics}, 42(4):763--808.

\bibitem[{Bowman et~al.(2015)Bowman, Angeli, Potts, and Manning}]{b21}
Samuel~R. Bowman, Gabor Angeli, Christopher Potts, and Christopher~D. Manning. 2015.
\newblock \href {https://doi.org/10.18653/v1/D15-1075} {A large annotated corpus for learning natural language inference}.
\newblock In \emph{Proceedings of the 2015 Conference on Empirical Methods in Natural Language Processing}, pages 632--642, Lisbon, Portugal. Association for Computational Linguistics.

\bibitem[{Brown et~al.(2020)Brown, Mann, Ryder, Subbiah, Kaplan, Dhariwal, Neelakantan, Shyam, Sastry, Askell, Agarwal, Herbert-Voss, Krueger, Henighan, Child, Ramesh, Ziegler, Wu, Winter, Hesse, Chen, Sigler, Litwin, Gray, Chess, Clark, Berner, McCandlish, Radford, Sutskever, and Amodei}]{brown2020language}
Tom~B Brown, Benjamin Mann, Nick Ryder, Melanie Subbiah, Jared Kaplan, Prafulla Dhariwal, Arvind Neelakantan, Pranav Shyam, Girish Sastry, Amanda Askell, Sandhini Agarwal, Ariel Herbert-Voss, Gretchen Krueger, Tom Henighan, Rewon Child, Aditya Ramesh, Daniel Ziegler, Jeffrey Wu, Clemens Winter, Chris Hesse, Mark Chen, Eric Sigler, Mateusz Litwin, Scott Gray, Benjamin Chess, Jack Clark, Christopher Berner, Sam McCandlish, Alec Radford, Ilya Sutskever, and Dario Amodei. 2020.
\newblock Language models are few-shot learners.
\newblock \emph{Advances in Neural Information Processing Systems}, 33:1877--1901.

\bibitem[{Bubeck et~al.(2023)Bubeck, Chandrasekaran, Eldan, Gehrke, Horvitz, Kamar, Lee, Lee, Li, Lundberg, Nori, Palangi, Ribeiro, and Zhang}]{bubeck2023sparks}
Sébastien Bubeck, Varun Chandrasekaran, Ronen Eldan, Johannes Gehrke, Eric Horvitz, Ece Kamar, Peter Lee, Yin~Tat Lee, Yuanzhi Li, Scott Lundberg, Harsha Nori, Hamid Palangi, Marco~Tulio Ribeiro, and Yi~Zhang. 2023.
\newblock \href {http://arxiv.org/abs/2303.12712} {Sparks of artificial general intelligence: Early experiments with gpt-4}.

\bibitem[{Chen et~al.(2024)Chen, Deng, Yuan, Ji, and Gu}]{chen2024selfplay}
Zixiang Chen, Yihe Deng, Huizhuo Yuan, Kaixuan Ji, and Quanquan Gu. 2024.
\newblock \href {http://arxiv.org/abs/2401.01335} {Self-play fine-tuning converts weak language models to strong language models}.

\bibitem[{Cosler et~al.(2023)Cosler, Hahn, Mendoza, Schmitt, and Trippel}]{b10}
Matthias Cosler, Christopher Hahn, Daniel Mendoza, Frederik Schmitt, and Caroline Trippel. 2023.
\newblock \href {https://doi.org/10.1007/978-3-031-37703-7\_18} {nl2spec: Interactively translating unstructured natural language to temporal logics with large language models}.
\newblock In \emph{Computer Aided Verification. CAV 2023. Lecture Notes in Computer Science}, volume 13965, Cham. Springer.

\bibitem[{Dalvi et~al.(2021)Dalvi, Jansen, Tafjord, Xie, Smith, Pipatanangkura, and Clark}]{dalvi2021explaining}
Bhavana Dalvi, Peter Jansen, Oyvind Tafjord, Zhengnan Xie, Hannah Smith, Leighanna Pipatanangkura, and Peter Clark. 2021.
\newblock \href {https://doi.org/10.18653/v1/2021.emnlp-main.585} {Explaining answers with entailment trees}.
\newblock In \emph{Proceedings of the 2021 Conference on Empirical Methods in Natural Language Processing}, pages 7358--7370, Online and Punta Cana, Dominican Republic. Association for Computational Linguistics.

\bibitem[{De~Moura and Bj\o{}rner(2008)}]{demoura2008z3}
Leonardo De~Moura and Nikolaj Bj\o{}rner. 2008.
\newblock Z3: an efficient smt solver.
\newblock In \emph{Proceedings of the Theory and Practice of Software, 14th International Conference on Tools and Algorithms for the Construction and Analysis of Systems}, TACAS'08/ETAPS'08, page 337–340, Berlin, Heidelberg. Springer-Verlag.

\bibitem[{Fuggitti and Chakraborti(2023)}]{b13}
Francesco Fuggitti and Tathagata Chakraborti. 2023.
\newblock \href {https://api.semanticscholar.org/CorpusID:259726762} {Nl2ltl - a python package for converting natural language (nl) instructions to linear temporal logic (ltl) formulas}.
\newblock In \emph{AAAI Conference on Artificial Intelligence}.

\bibitem[{Gendron et~al.(2024)Gendron, Bao, Witbrock, and Dobbie}]{gendron2024largelanguagemodelsstrong}
Gaël Gendron, Qiming Bao, Michael Witbrock, and Gillian Dobbie. 2024.
\newblock \href {http://arxiv.org/abs/2305.19555} {Large language models are not strong abstract reasoners}.

\bibitem[{Grattafiori et~al.(2024)Grattafiori, Dubey, Jauhri, Pandey, Kadian, Al-Dahle, Letman, Mathur, Schelten, Vaughan, Yang, Fan, Goyal, Hartshorn, Yang, Mitra, Sravankumar, Korenev, Hinsvark, Rao, Zhang, Rodriguez, Gregerson, Spataru, Roziere, Biron, Tang, Chern, Caucheteux, Nayak, Bi, Marra, McConnell, Keller, Touret, Wu, Wong, Ferrer, Nikolaidis, Allonsius, Song, Pintz, Livshits, Wyatt, Esiobu, Choudhary, Mahajan, Garcia-Olano, Perino, Hupkes, Lakomkin, AlBadawy, Lobanova, Dinan, Smith, Radenovic, Guzmán, Zhang, Synnaeve, Lee, Anderson, Thattai, Nail, Mialon, Pang, Cucurell, Nguyen, Korevaar, Xu, Touvron, Zarov, Ibarra, Kloumann, Misra, Evtimov, Zhang, Copet, Lee, Geffert, Vranes, Park, Mahadeokar, Shah, van~der Linde, Billock, Hong, Lee, Fu, Chi, Huang, Liu, Wang, Yu, Bitton, Spisak, Park, Rocca, Johnstun, Saxe, Jia, Alwala, Prasad, Upasani, Plawiak, Li, Heafield, Stone, El-Arini, Iyer, Malik, Chiu, Bhalla, Lakhotia, Rantala-Yeary, van~der Maaten, Chen, Tan, Jenkins, Martin, Madaan, Malo, Blecher,
  Landzaat, de~Oliveira, Muzzi, Pasupuleti, Singh, Paluri, Kardas, Tsimpoukelli, Oldham, Rita, Pavlova, Kambadur, Lewis, Si, Singh, Hassan, Goyal, Torabi, Bashlykov, Bogoychev, Chatterji, Zhang, Duchenne, Çelebi, Alrassy, Zhang, Li, Vasic, Weng, Bhargava, Dubal, Krishnan, Koura, Xu, He, Dong, Srinivasan, Ganapathy, Calderer, Cabral, Stojnic, Raileanu, Maheswari, Girdhar, Patel, Sauvestre, Polidoro, Sumbaly, Taylor, Silva, Hou, Wang, Hosseini, Chennabasappa, Singh, Bell, Kim, Edunov, Nie, Narang, Raparthy, Shen, Wan, Bhosale, Zhang, Vandenhende, Batra, Whitman, Sootla, Collot, Gururangan, Borodinsky, Herman, Fowler, Sheasha, Georgiou, Scialom, Speckbacher, Mihaylov, Xiao, Karn, Goswami, Gupta, Ramanathan, Kerkez, Gonguet, Do, Vogeti, Albiero, Petrovic, Chu, Xiong, Fu, Meers, Martinet, Wang, Wang, Tan, Xia, Xie, Jia, Wang, Goldschlag, Gaur, Babaei, Wen, Song, Zhang, Li, Mao, Coudert, Yan, Chen, Papakipos, Singh, Srivastava, Jain, Kelsey, Shajnfeld, Gangidi, Victoria, Goldstand, Menon, Sharma, Boesenberg,
  Baevski, Feinstein, Kallet, Sangani, Teo, Yunus, Lupu, Alvarado, Caples, Gu, Ho, Poulton, Ryan, Ramchandani, Dong, Franco, Goyal, Saraf, Chowdhury, Gabriel, Bharambe, Eisenman, Yazdan, James, Maurer, Leonhardi, Huang, Loyd, Paola, Paranjape, Liu, Wu, Ni, Hancock, Wasti, Spence, Stojkovic, Gamido, Montalvo, Parker, Burton, Mejia, Liu, Wang, Kim, Zhou, Hu, Chu, Cai, Tindal, Feichtenhofer, Gao, Civin, Beaty, Kreymer, Li, Adkins, Xu, Testuggine, David, Parikh, Liskovich, Foss, Wang, Le, Holland, Dowling, Jamil, Montgomery, Presani, Hahn, Wood, Le, Brinkman, Arcaute, Dunbar, Smothers, Sun, Kreuk, Tian, Kokkinos, Ozgenel, Caggioni, Kanayet, Seide, Florez, Schwarz, Badeer, Swee, Halpern, Herman, Sizov, Guangyi, Zhang, Lakshminarayanan, Inan, Shojanazeri, Zou, Wang, Zha, Habeeb, Rudolph, Suk, Aspegren, Goldman, Zhan, Damlaj, Molybog, Tufanov, Leontiadis, Veliche, Gat, Weissman, Geboski, Kohli, Lam, Asher, Gaya, Marcus, Tang, Chan, Zhen, Reizenstein, Teboul, Zhong, Jin, Yang, Cummings, Carvill, Shepard, McPhie,
  Torres, Ginsburg, Wang, Wu, U, Saxena, Khandelwal, Zand, Matosich, Veeraraghavan, Michelena, Li, Jagadeesh, Huang, Chawla, Huang, Chen, Garg, A, Silva, Bell, Zhang, Guo, Yu, Moshkovich, Wehrstedt, Khabsa, Avalani, Bhatt, Mankus, Hasson, Lennie, Reso, Groshev, Naumov, Lathi, Keneally, Liu, Seltzer, Valko, Restrepo, Patel, Vyatskov, Samvelyan, Clark, Macey, Wang, Hermoso, Metanat, Rastegari, Bansal, Santhanam, Parks, White, Bawa, Singhal, Egebo, Usunier, Mehta, Laptev, Dong, Cheng, Chernoguz, Hart, Salpekar, Kalinli, Kent, Parekh, Saab, Balaji, Rittner, Bontrager, Roux, Dollar, Zvyagina, Ratanchandani, Yuvraj, Liang, Alao, Rodriguez, Ayub, Murthy, Nayani, Mitra, Parthasarathy, Li, Hogan, Battey, Wang, Howes, Rinott, Mehta, Siby, Bondu, Datta, Chugh, Hunt, Dhillon, Sidorov, Pan, Mahajan, Verma, Yamamoto, Ramaswamy, Lindsay, Lindsay, Feng, Lin, Zha, Patil, Shankar, Zhang, Zhang, Wang, Agarwal, Sajuyigbe, Chintala, Max, Chen, Kehoe, Satterfield, Govindaprasad, Gupta, Deng, Cho, Virk, Subramanian, Choudhury,
  Goldman, Remez, Glaser, Best, Koehler, Robinson, Li, Zhang, Matthews, Chou, Shaked, Vontimitta, Ajayi, Montanez, Mohan, Kumar, Mangla, Ionescu, Poenaru, Mihailescu, Ivanov, Li, Wang, Jiang, Bouaziz, Constable, Tang, Wu, Wang, Wu, Gao, Kleinman, Chen, Hu, Jia, Qi, Li, Zhang, Zhang, Adi, Nam, Yu, Wang, Zhao, Hao, Qian, Li, He, Rait, DeVito, Rosnbrick, Wen, Yang, Zhao, and Ma}]{grattafiori2024llama3herdmodels}
Aaron Grattafiori, Abhimanyu Dubey, Abhinav Jauhri, Abhinav Pandey, Abhishek Kadian, Ahmad Al-Dahle, Aiesha Letman, Akhil Mathur, Alan Schelten, Alex Vaughan, Amy Yang, Angela Fan, Anirudh Goyal, Anthony Hartshorn, Aobo Yang, Archi Mitra, Archie Sravankumar, Artem Korenev, Arthur Hinsvark, Arun Rao, Aston Zhang, Aurelien Rodriguez, Austen Gregerson, Ava Spataru, Baptiste Roziere, Bethany Biron, Binh Tang, Bobbie Chern, Charlotte Caucheteux, Chaya Nayak, Chloe Bi, Chris Marra, Chris McConnell, Christian Keller, Christophe Touret, Chunyang Wu, Corinne Wong, Cristian~Canton Ferrer, Cyrus Nikolaidis, Damien Allonsius, Daniel Song, Danielle Pintz, Danny Livshits, Danny Wyatt, David Esiobu, Dhruv Choudhary, Dhruv Mahajan, Diego Garcia-Olano, Diego Perino, Dieuwke Hupkes, Egor Lakomkin, Ehab AlBadawy, Elina Lobanova, Emily Dinan, Eric~Michael Smith, Filip Radenovic, Francisco Guzmán, Frank Zhang, Gabriel Synnaeve, Gabrielle Lee, Georgia~Lewis Anderson, Govind Thattai, Graeme Nail, Gregoire Mialon, Guan Pang,
  Guillem Cucurell, Hailey Nguyen, Hannah Korevaar, Hu~Xu, Hugo Touvron, Iliyan Zarov, Imanol~Arrieta Ibarra, Isabel Kloumann, Ishan Misra, Ivan Evtimov, Jack Zhang, Jade Copet, Jaewon Lee, Jan Geffert, Jana Vranes, Jason Park, Jay Mahadeokar, Jeet Shah, Jelmer van~der Linde, Jennifer Billock, Jenny Hong, Jenya Lee, Jeremy Fu, Jianfeng Chi, Jianyu Huang, Jiawen Liu, Jie Wang, Jiecao Yu, Joanna Bitton, Joe Spisak, Jongsoo Park, Joseph Rocca, Joshua Johnstun, Joshua Saxe, Junteng Jia, Kalyan~Vasuden Alwala, Karthik Prasad, Kartikeya Upasani, Kate Plawiak, Ke~Li, Kenneth Heafield, Kevin Stone, Khalid El-Arini, Krithika Iyer, Kshitiz Malik, Kuenley Chiu, Kunal Bhalla, Kushal Lakhotia, Lauren Rantala-Yeary, Laurens van~der Maaten, Lawrence Chen, Liang Tan, Liz Jenkins, Louis Martin, Lovish Madaan, Lubo Malo, Lukas Blecher, Lukas Landzaat, Luke de~Oliveira, Madeline Muzzi, Mahesh Pasupuleti, Mannat Singh, Manohar Paluri, Marcin Kardas, Maria Tsimpoukelli, Mathew Oldham, Mathieu Rita, Maya Pavlova, Melanie Kambadur,
  Mike Lewis, Min Si, Mitesh~Kumar Singh, Mona Hassan, Naman Goyal, Narjes Torabi, Nikolay Bashlykov, Nikolay Bogoychev, Niladri Chatterji, Ning Zhang, Olivier Duchenne, Onur Çelebi, Patrick Alrassy, Pengchuan Zhang, Pengwei Li, Petar Vasic, Peter Weng, Prajjwal Bhargava, Pratik Dubal, Praveen Krishnan, Punit~Singh Koura, Puxin Xu, Qing He, Qingxiao Dong, Ragavan Srinivasan, Raj Ganapathy, Ramon Calderer, Ricardo~Silveira Cabral, Robert Stojnic, Roberta Raileanu, Rohan Maheswari, Rohit Girdhar, Rohit Patel, Romain Sauvestre, Ronnie Polidoro, Roshan Sumbaly, Ross Taylor, Ruan Silva, Rui Hou, Rui Wang, Saghar Hosseini, Sahana Chennabasappa, Sanjay Singh, Sean Bell, Seohyun~Sonia Kim, Sergey Edunov, Shaoliang Nie, Sharan Narang, Sharath Raparthy, Sheng Shen, Shengye Wan, Shruti Bhosale, Shun Zhang, Simon Vandenhende, Soumya Batra, Spencer Whitman, Sten Sootla, Stephane Collot, Suchin Gururangan, Sydney Borodinsky, Tamar Herman, Tara Fowler, Tarek Sheasha, Thomas Georgiou, Thomas Scialom, Tobias Speckbacher,
  Todor Mihaylov, Tong Xiao, Ujjwal Karn, Vedanuj Goswami, Vibhor Gupta, Vignesh Ramanathan, Viktor Kerkez, Vincent Gonguet, Virginie Do, Vish Vogeti, Vítor Albiero, Vladan Petrovic, Weiwei Chu, Wenhan Xiong, Wenyin Fu, Whitney Meers, Xavier Martinet, Xiaodong Wang, Xiaofang Wang, Xiaoqing~Ellen Tan, Xide Xia, Xinfeng Xie, Xuchao Jia, Xuewei Wang, Yaelle Goldschlag, Yashesh Gaur, Yasmine Babaei, Yi~Wen, Yiwen Song, Yuchen Zhang, Yue Li, Yuning Mao, Zacharie~Delpierre Coudert, Zheng Yan, Zhengxing Chen, Zoe Papakipos, Aaditya Singh, Aayushi Srivastava, Abha Jain, Adam Kelsey, Adam Shajnfeld, Adithya Gangidi, Adolfo Victoria, Ahuva Goldstand, Ajay Menon, Ajay Sharma, Alex Boesenberg, Alexei Baevski, Allie Feinstein, Amanda Kallet, Amit Sangani, Amos Teo, Anam Yunus, Andrei Lupu, Andres Alvarado, Andrew Caples, Andrew Gu, Andrew Ho, Andrew Poulton, Andrew Ryan, Ankit Ramchandani, Annie Dong, Annie Franco, Anuj Goyal, Aparajita Saraf, Arkabandhu Chowdhury, Ashley Gabriel, Ashwin Bharambe, Assaf Eisenman, Azadeh
  Yazdan, Beau James, Ben Maurer, Benjamin Leonhardi, Bernie Huang, Beth Loyd, Beto~De Paola, Bhargavi Paranjape, Bing Liu, Bo~Wu, Boyu Ni, Braden Hancock, Bram Wasti, Brandon Spence, Brani Stojkovic, Brian Gamido, Britt Montalvo, Carl Parker, Carly Burton, Catalina Mejia, Ce~Liu, Changhan Wang, Changkyu Kim, Chao Zhou, Chester Hu, Ching-Hsiang Chu, Chris Cai, Chris Tindal, Christoph Feichtenhofer, Cynthia Gao, Damon Civin, Dana Beaty, Daniel Kreymer, Daniel Li, David Adkins, David Xu, Davide Testuggine, Delia David, Devi Parikh, Diana Liskovich, Didem Foss, Dingkang Wang, Duc Le, Dustin Holland, Edward Dowling, Eissa Jamil, Elaine Montgomery, Eleonora Presani, Emily Hahn, Emily Wood, Eric-Tuan Le, Erik Brinkman, Esteban Arcaute, Evan Dunbar, Evan Smothers, Fei Sun, Felix Kreuk, Feng Tian, Filippos Kokkinos, Firat Ozgenel, Francesco Caggioni, Frank Kanayet, Frank Seide, Gabriela~Medina Florez, Gabriella Schwarz, Gada Badeer, Georgia Swee, Gil Halpern, Grant Herman, Grigory Sizov, Guangyi, Zhang, Guna
  Lakshminarayanan, Hakan Inan, Hamid Shojanazeri, Han Zou, Hannah Wang, Hanwen Zha, Haroun Habeeb, Harrison Rudolph, Helen Suk, Henry Aspegren, Hunter Goldman, Hongyuan Zhan, Ibrahim Damlaj, Igor Molybog, Igor Tufanov, Ilias Leontiadis, Irina-Elena Veliche, Itai Gat, Jake Weissman, James Geboski, James Kohli, Janice Lam, Japhet Asher, Jean-Baptiste Gaya, Jeff Marcus, Jeff Tang, Jennifer Chan, Jenny Zhen, Jeremy Reizenstein, Jeremy Teboul, Jessica Zhong, Jian Jin, Jingyi Yang, Joe Cummings, Jon Carvill, Jon Shepard, Jonathan McPhie, Jonathan Torres, Josh Ginsburg, Junjie Wang, Kai Wu, Kam~Hou U, Karan Saxena, Kartikay Khandelwal, Katayoun Zand, Kathy Matosich, Kaushik Veeraraghavan, Kelly Michelena, Keqian Li, Kiran Jagadeesh, Kun Huang, Kunal Chawla, Kyle Huang, Lailin Chen, Lakshya Garg, Lavender A, Leandro Silva, Lee Bell, Lei Zhang, Liangpeng Guo, Licheng Yu, Liron Moshkovich, Luca Wehrstedt, Madian Khabsa, Manav Avalani, Manish Bhatt, Martynas Mankus, Matan Hasson, Matthew Lennie, Matthias Reso, Maxim
  Groshev, Maxim Naumov, Maya Lathi, Meghan Keneally, Miao Liu, Michael~L. Seltzer, Michal Valko, Michelle Restrepo, Mihir Patel, Mik Vyatskov, Mikayel Samvelyan, Mike Clark, Mike Macey, Mike Wang, Miquel~Jubert Hermoso, Mo~Metanat, Mohammad Rastegari, Munish Bansal, Nandhini Santhanam, Natascha Parks, Natasha White, Navyata Bawa, Nayan Singhal, Nick Egebo, Nicolas Usunier, Nikhil Mehta, Nikolay~Pavlovich Laptev, Ning Dong, Norman Cheng, Oleg Chernoguz, Olivia Hart, Omkar Salpekar, Ozlem Kalinli, Parkin Kent, Parth Parekh, Paul Saab, Pavan Balaji, Pedro Rittner, Philip Bontrager, Pierre Roux, Piotr Dollar, Polina Zvyagina, Prashant Ratanchandani, Pritish Yuvraj, Qian Liang, Rachad Alao, Rachel Rodriguez, Rafi Ayub, Raghotham Murthy, Raghu Nayani, Rahul Mitra, Rangaprabhu Parthasarathy, Raymond Li, Rebekkah Hogan, Robin Battey, Rocky Wang, Russ Howes, Ruty Rinott, Sachin Mehta, Sachin Siby, Sai~Jayesh Bondu, Samyak Datta, Sara Chugh, Sara Hunt, Sargun Dhillon, Sasha Sidorov, Satadru Pan, Saurabh Mahajan,
  Saurabh Verma, Seiji Yamamoto, Sharadh Ramaswamy, Shaun Lindsay, Shaun Lindsay, Sheng Feng, Shenghao Lin, Shengxin~Cindy Zha, Shishir Patil, Shiva Shankar, Shuqiang Zhang, Shuqiang Zhang, Sinong Wang, Sneha Agarwal, Soji Sajuyigbe, Soumith Chintala, Stephanie Max, Stephen Chen, Steve Kehoe, Steve Satterfield, Sudarshan Govindaprasad, Sumit Gupta, Summer Deng, Sungmin Cho, Sunny Virk, Suraj Subramanian, Sy~Choudhury, Sydney Goldman, Tal Remez, Tamar Glaser, Tamara Best, Thilo Koehler, Thomas Robinson, Tianhe Li, Tianjun Zhang, Tim Matthews, Timothy Chou, Tzook Shaked, Varun Vontimitta, Victoria Ajayi, Victoria Montanez, Vijai Mohan, Vinay~Satish Kumar, Vishal Mangla, Vlad Ionescu, Vlad Poenaru, Vlad~Tiberiu Mihailescu, Vladimir Ivanov, Wei Li, Wenchen Wang, Wenwen Jiang, Wes Bouaziz, Will Constable, Xiaocheng Tang, Xiaojian Wu, Xiaolan Wang, Xilun Wu, Xinbo Gao, Yaniv Kleinman, Yanjun Chen, Ye~Hu, Ye~Jia, Ye~Qi, Yenda Li, Yilin Zhang, Ying Zhang, Yossi Adi, Youngjin Nam, Yu, Wang, Yu~Zhao, Yuchen Hao, Yundi
  Qian, Yunlu Li, Yuzi He, Zach Rait, Zachary DeVito, Zef Rosnbrick, Zhaoduo Wen, Zhenyu Yang, Zhiwei Zhao, and Zhiyu Ma. 2024.
\newblock \href {http://arxiv.org/abs/2407.21783} {The llama 3 herd of models}.

\bibitem[{Habernal et~al.(2018{\natexlab{a}})Habernal, Wachsmuth, Gurevych, and Kiesel}]{b4}
Ivan Habernal, Henning Wachsmuth, Iryna Gurevych, and Benno Kiesel. 2018{\natexlab{a}}.
\newblock "dummy, grandpa, do you know anything?": Identifying and characterizing ad hominem fallacies in the wild.
\newblock In \emph{Proceedings of the 12th International AAAI Conference on Web and Social Media (ICWSM)}, pages 206--215.

\bibitem[{Habernal et~al.(2018{\natexlab{b}})Habernal, Wachsmuth, Gurevych, and Stein}]{b3}
Ivan Habernal, Henning Wachsmuth, Iryna Gurevych, and Benno Stein. 2018{\natexlab{b}}.
\newblock \href {https://doi.org/10.18653/v1/N18-1036} {Before name-calling: Dynamics and triggers of ad hominem fallacies in web argumentation}.
\newblock In \emph{Proceedings of the 2018 Conference of the North {A}merican Chapter of the Association for Computational Linguistics: Human Language Technologies, Volume 1 (Long Papers)}, pages 386--396, New Orleans, Louisiana. Association for Computational Linguistics.

\bibitem[{Hahn et~al.(2022)Hahn, Schmitt, Tillman, Metzger, Siber, and Finkbeiner}]{hahn2022formal}
Christopher Hahn, Frederik Schmitt, Julia~J. Tillman, Niklas Metzger, Julian Siber, and Bernd Finkbeiner. 2022.
\newblock \href {http://arxiv.org/abs/2206.01962} {Formal specifications from natural language}.

\bibitem[{Haluptzok et~al.(2022)Haluptzok, Bowers, and Kalai}]{jin2023iterative}
Patrick~M. Haluptzok, Matthew Bowers, and Adam~Tauman Kalai. 2022.
\newblock \href {https://api.semanticscholar.org/CorpusID:251197051} {Language modexrls can teach themselves to program better}.
\newblock \emph{ArXiv}, abs/2207.14502.

\bibitem[{Jin et~al.(2022)Jin, Lalwani, Vaidhya, Shen, Ding, Lyu, Sachan, Mihalcea, and Schoelkopf}]{jin2022logical}
Zhijing Jin, Abhinav Lalwani, Tejas Vaidhya, Xiaoyu Shen, Yiwen Ding, Zhiheng Lyu, Mrinmaya Sachan, Rada Mihalcea, and Bernhard Schoelkopf. 2022.
\newblock \href {https://doi.org/10.18653/v1/2022.findings-emnlp.532} {Logical fallacy detection}.
\newblock In \emph{Findings of the Association for Computational Linguistics: EMNLP 2022}, pages 7180--7198, Abu Dhabi, United Arab Emirates. Association for Computational Linguistics.

\bibitem[{Lewis et~al.(2020)Lewis, Liu, Goyal, Ghazvininejad, Mohamed, Levy, Stoyanov, and Zettlemoyer}]{b20}
Mike Lewis, Yinhan Liu, Naman Goyal, Marjan Ghazvininejad, Abdelrahman Mohamed, Omer Levy, Veselin Stoyanov, and Luke Zettlemoyer. 2020.
\newblock \href {https://doi.org/10.18653/v1/2020.acl-main.703} {{BART}: Denoising sequence-to-sequence pre-training for natural language generation, translation, and comprehension}.
\newblock In \emph{Proceedings of the 58th Annual Meeting of the Association for Computational Linguistics}, pages 7871--7880, Online. Association for Computational Linguistics.

\bibitem[{Liu et~al.(2022)Liu, Yang, Schornstein, Liang, Idrees, Tellex, and Shah}]{b14}
Jason~Xinyu Liu, Ziyi Yang, Benjamin Schornstein, Sam Liang, Ifrah Idrees, Stefanie Tellex, and Ankit Shah. 2022.
\newblock {Lang2LTL}: Translating natural language commands to temporal specification with large language models.
\newblock In \emph{CoRL Workshop on Language and Robot Learning}.

\bibitem[{MacCartney and Manning(2014)}]{b32}
B.~MacCartney and C.~D. Manning. 2014.
\newblock \href {https://doi.org/10.1007/978-94-007-7284-7_8} {Natural logic and natural language inference}.
\newblock In H.~Bunt, J.~Bos, and S.~Pulman, editors, \emph{Computing Meaning: Volume 4}, pages 129--147. Springer Netherlands, Dordrecht.

\bibitem[{Nakpih and Santini(2020)}]{b5}
Callistus~Ireneous Nakpih and Simone Santini. 2020.
\newblock \href {https://api.semanticscholar.org/CorpusID:214718344} {Automated discovery of logical fallacies in legal argumentation}.
\newblock \emph{International Journal of Artificial Intelligence \& Applications}.

\bibitem[{Olausson et~al.(2023)Olausson, Gu, Lipkin, Zhang, Solar-Lezama, Tenenbaum, and Levy}]{olausson-etal-2023-linc}
Theo Olausson, Alex Gu, Ben Lipkin, Cedegao Zhang, Armando Solar-Lezama, Joshua Tenenbaum, and Roger Levy. 2023.
\newblock \href {https://doi.org/10.18653/v1/2023.emnlp-main.313} {{LINC}: A neurosymbolic approach for logical reasoning by combining language models with first-order logic provers}.
\newblock In \emph{Proceedings of the 2023 Conference on Empirical Methods in Natural Language Processing}, pages 5153--5176, Singapore. Association for Computational Linguistics.

\bibitem[{OpenAI(2024)}]{4omini}
OpenAI. 2024.
\newblock \href {https://openai.com/index/gpt-4o-mini-advancing-cost-efficient-intelligence/} {Gpt-4o mini: Advancing cost-efficient intelligence}.
\newblock Accessed: 2025-02-15.

\bibitem[{OpenAI et~al.(2024{\natexlab{a}})OpenAI, Hurst, Lerer, Goucher, Perelman, Ramesh, Clark, Ostrow, Welihinda, Hayes, Radford, Mądry, Baker-Whitcomb, Beutel, Borzunov, Carney, Chow, Kirillov, Nichol, Paino, Renzin, Passos, Kirillov, Christakis, Conneau, Kamali, Jabri, Moyer, Tam, Crookes, Tootoochian, Tootoonchian, Kumar, Vallone, Karpathy, Braunstein, Cann, Codispoti, Galu, Kondrich, Tulloch, Mishchenko, Baek, Jiang, Pelisse, Woodford, Gosalia, Dhar, Pantuliano, Nayak, Oliver, Zoph, Ghorbani, Leimberger, Rossen, Sokolowsky, Wang, Zweig, Hoover, Samic, McGrew, Spero, Giertler, Cheng, Lightcap, Walkin, Quinn, Guarraci, Hsu, Kellogg, Eastman, Lugaresi, Wainwright, Bassin, Hudson, Chu, Nelson, Li, Shern, Conger, Barette, Voss, Ding, Lu, Zhang, Beaumont, Hallacy, Koch, Gibson, Kim, Choi, McLeavey, Hesse, Fischer, Winter, Czarnecki, Jarvis, Wei, Koumouzelis, Sherburn, Kappler, Levin, Levy, Carr, Farhi, Mely, Robinson, Sasaki, Jin, Valladares, Tsipras, Li, Nguyen, Findlay, Oiwoh, Wong, Asdar, Proehl, Yang,
  Antonow, Kramer, Peterson, Sigler, Wallace, Brevdo, Mays, Khorasani, Such, Raso, Zhang, von Lohmann, Sulit, Goh, Oden, Salmon, Starace, Brockman, Salman, Bao, Hu, Wong, Wang, Schmidt, Whitney, Jun, Kirchner, de~Oliveira~Pinto, Ren, Chang, Chung, Kivlichan, O'Connell, O'Connell, Osband, Silber, Sohl, Okuyucu, Lan, Kostrikov, Sutskever, Kanitscheider, Gulrajani, Coxon, Menick, Pachocki, Aung, Betker, Crooks, Lennon, Kiros, Leike, Park, Kwon, Phang, Teplitz, Wei, Wolfe, Chen, Harris, Varavva, Lee, Shieh, Lin, Yu, Weng, Tang, Yu, Jang, Candela, Beutler, Landers, Parish, Heidecke, Schulman, Lachman, McKay, Uesato, Ward, Kim, Huizinga, Sitkin, Kraaijeveld, Gross, Kaplan, Snyder, Achiam, Jiao, Lee, Zhuang, Harriman, Fricke, Hayashi, Singhal, Shi, Karthik, Wood, Rimbach, Hsu, Nguyen, Gu-Lemberg, Button, Liu, Howe, Muthukumar, Luther, Ahmad, Kai, Itow, Workman, Pathak, Chen, Jing, Guy, Fedus, Zhou, Mamitsuka, Weng, McCallum, Held, Ouyang, Feuvrier, Zhang, Kondraciuk, Kaiser, Hewitt, Metz, Doshi, Aflak, Simens, Boyd,
  Thompson, Dukhan, Chen, Gray, Hudnall, Zhang, Aljubeh, Litwin, Zeng, Johnson, Shetty, Gupta, Shah, Yatbaz, Yang, Zhong, Glaese, Chen, Janner, Lampe, Petrov, Wu, Wang, Fradin, Pokrass, Castro, de~Castro, Pavlov, Brundage, Wang, Khan, Murati, Bavarian, Lin, Yesildal, Soto, Gimelshein, Cone, Staudacher, Summers, LaFontaine, Chowdhury, Ryder, Stathas, Turley, Tezak, Felix, Kudige, Keskar, Deutsch, Bundick, Puckett, Nachum, Okelola, Boiko, Murk, Jaffe, Watkins, Godement, Campbell-Moore, Chao, McMillan, Belov, Su, Bak, Bakkum, Deng, Dolan, Hoeschele, Welinder, Tillet, Pronin, Tillet, Dhariwal, Yuan, Dias, Lim, Arora, Troll, Lin, Lopes, Puri, Miyara, Leike, Gaubert, Zamani, Wang, Donnelly, Honsby, Smith, Sahai, Ramchandani, Huet, Carmichael, Zellers, Chen, Chen, Nigmatullin, Cheu, Jain, Altman, Schoenholz, Toizer, Miserendino, Agarwal, Culver, Ethersmith, Gray, Grove, Metzger, Hermani, Jain, Zhao, Wu, Jomoto, Wu, Shuaiqi, Xia, Phene, Papay, Narayanan, Coffey, Lee, Hall, Balaji, Broda, Stramer, Xu, Gogineni,
  Christianson, Sanders, Patwardhan, Cunninghman, Degry, Dimson, Raoux, Shadwell, Zheng, Underwood, Markov, Sherbakov, Rubin, Stasi, Kaftan, Heywood, Peterson, Walters, Eloundou, Qi, Moeller, Monaco, Kuo, Fomenko, Chang, Zheng, Zhou, Manassra, Sheu, Zaremba, Patil, Qian, Kim, Cheng, Zhang, He, Zhang, Jin, Dai, and Malkov}]{openai2024gpt4ocard}
OpenAI, Aaron Hurst, Adam Lerer, Adam~P. Goucher, Adam Perelman, Aditya Ramesh, Aidan Clark, AJ~Ostrow, Akila Welihinda, Alan Hayes, Alec Radford, Aleksander Mądry, Alex Baker-Whitcomb, Alex Beutel, Alex Borzunov, Alex Carney, Alex Chow, Alex Kirillov, Alex Nichol, Alex Paino, Alex Renzin, Alex~Tachard Passos, Alexander Kirillov, Alexi Christakis, Alexis Conneau, Ali Kamali, Allan Jabri, Allison Moyer, Allison Tam, Amadou Crookes, Amin Tootoochian, Amin Tootoonchian, Ananya Kumar, Andrea Vallone, Andrej Karpathy, Andrew Braunstein, Andrew Cann, Andrew Codispoti, Andrew Galu, Andrew Kondrich, Andrew Tulloch, Andrey Mishchenko, Angela Baek, Angela Jiang, Antoine Pelisse, Antonia Woodford, Anuj Gosalia, Arka Dhar, Ashley Pantuliano, Avi Nayak, Avital Oliver, Barret Zoph, Behrooz Ghorbani, Ben Leimberger, Ben Rossen, Ben Sokolowsky, Ben Wang, Benjamin Zweig, Beth Hoover, Blake Samic, Bob McGrew, Bobby Spero, Bogo Giertler, Bowen Cheng, Brad Lightcap, Brandon Walkin, Brendan Quinn, Brian Guarraci, Brian Hsu,
  Bright Kellogg, Brydon Eastman, Camillo Lugaresi, Carroll Wainwright, Cary Bassin, Cary Hudson, Casey Chu, Chad Nelson, Chak Li, Chan~Jun Shern, Channing Conger, Charlotte Barette, Chelsea Voss, Chen Ding, Cheng Lu, Chong Zhang, Chris Beaumont, Chris Hallacy, Chris Koch, Christian Gibson, Christina Kim, Christine Choi, Christine McLeavey, Christopher Hesse, Claudia Fischer, Clemens Winter, Coley Czarnecki, Colin Jarvis, Colin Wei, Constantin Koumouzelis, Dane Sherburn, Daniel Kappler, Daniel Levin, Daniel Levy, David Carr, David Farhi, David Mely, David Robinson, David Sasaki, Denny Jin, Dev Valladares, Dimitris Tsipras, Doug Li, Duc~Phong Nguyen, Duncan Findlay, Edede Oiwoh, Edmund Wong, Ehsan Asdar, Elizabeth Proehl, Elizabeth Yang, Eric Antonow, Eric Kramer, Eric Peterson, Eric Sigler, Eric Wallace, Eugene Brevdo, Evan Mays, Farzad Khorasani, Felipe~Petroski Such, Filippo Raso, Francis Zhang, Fred von Lohmann, Freddie Sulit, Gabriel Goh, Gene Oden, Geoff Salmon, Giulio Starace, Greg Brockman, Hadi
  Salman, Haiming Bao, Haitang Hu, Hannah Wong, Haoyu Wang, Heather Schmidt, Heather Whitney, Heewoo Jun, Hendrik Kirchner, Henrique~Ponde de~Oliveira~Pinto, Hongyu Ren, Huiwen Chang, Hyung~Won Chung, Ian Kivlichan, Ian O'Connell, Ian O'Connell, Ian Osband, Ian Silber, Ian Sohl, Ibrahim Okuyucu, Ikai Lan, Ilya Kostrikov, Ilya Sutskever, Ingmar Kanitscheider, Ishaan Gulrajani, Jacob Coxon, Jacob Menick, Jakub Pachocki, James Aung, James Betker, James Crooks, James Lennon, Jamie Kiros, Jan Leike, Jane Park, Jason Kwon, Jason Phang, Jason Teplitz, Jason Wei, Jason Wolfe, Jay Chen, Jeff Harris, Jenia Varavva, Jessica~Gan Lee, Jessica Shieh, Ji~Lin, Jiahui Yu, Jiayi Weng, Jie Tang, Jieqi Yu, Joanne Jang, Joaquin~Quinonero Candela, Joe Beutler, Joe Landers, Joel Parish, Johannes Heidecke, John Schulman, Jonathan Lachman, Jonathan McKay, Jonathan Uesato, Jonathan Ward, Jong~Wook Kim, Joost Huizinga, Jordan Sitkin, Jos Kraaijeveld, Josh Gross, Josh Kaplan, Josh Snyder, Joshua Achiam, Joy Jiao, Joyce Lee, Juntang
  Zhuang, Justyn Harriman, Kai Fricke, Kai Hayashi, Karan Singhal, Katy Shi, Kavin Karthik, Kayla Wood, Kendra Rimbach, Kenny Hsu, Kenny Nguyen, Keren Gu-Lemberg, Kevin Button, Kevin Liu, Kiel Howe, Krithika Muthukumar, Kyle Luther, Lama Ahmad, Larry Kai, Lauren Itow, Lauren Workman, Leher Pathak, Leo Chen, Li~Jing, Lia Guy, Liam Fedus, Liang Zhou, Lien Mamitsuka, Lilian Weng, Lindsay McCallum, Lindsey Held, Long Ouyang, Louis Feuvrier, Lu~Zhang, Lukas Kondraciuk, Lukasz Kaiser, Luke Hewitt, Luke Metz, Lyric Doshi, Mada Aflak, Maddie Simens, Madelaine Boyd, Madeleine Thompson, Marat Dukhan, Mark Chen, Mark Gray, Mark Hudnall, Marvin Zhang, Marwan Aljubeh, Mateusz Litwin, Matthew Zeng, Max Johnson, Maya Shetty, Mayank Gupta, Meghan Shah, Mehmet Yatbaz, Meng~Jia Yang, Mengchao Zhong, Mia Glaese, Mianna Chen, Michael Janner, Michael Lampe, Michael Petrov, Michael Wu, Michele Wang, Michelle Fradin, Michelle Pokrass, Miguel Castro, Miguel Oom~Temudo de~Castro, Mikhail Pavlov, Miles Brundage, Miles Wang, Minal
  Khan, Mira Murati, Mo~Bavarian, Molly Lin, Murat Yesildal, Nacho Soto, Natalia Gimelshein, Natalie Cone, Natalie Staudacher, Natalie Summers, Natan LaFontaine, Neil Chowdhury, Nick Ryder, Nick Stathas, Nick Turley, Nik Tezak, Niko Felix, Nithanth Kudige, Nitish Keskar, Noah Deutsch, Noel Bundick, Nora Puckett, Ofir Nachum, Ola Okelola, Oleg Boiko, Oleg Murk, Oliver Jaffe, Olivia Watkins, Olivier Godement, Owen Campbell-Moore, Patrick Chao, Paul McMillan, Pavel Belov, Peng Su, Peter Bak, Peter Bakkum, Peter Deng, Peter Dolan, Peter Hoeschele, Peter Welinder, Phil Tillet, Philip Pronin, Philippe Tillet, Prafulla Dhariwal, Qiming Yuan, Rachel Dias, Rachel Lim, Rahul Arora, Rajan Troll, Randall Lin, Rapha~Gontijo Lopes, Raul Puri, Reah Miyara, Reimar Leike, Renaud Gaubert, Reza Zamani, Ricky Wang, Rob Donnelly, Rob Honsby, Rocky Smith, Rohan Sahai, Rohit Ramchandani, Romain Huet, Rory Carmichael, Rowan Zellers, Roy Chen, Ruby Chen, Ruslan Nigmatullin, Ryan Cheu, Saachi Jain, Sam Altman, Sam Schoenholz, Sam
  Toizer, Samuel Miserendino, Sandhini Agarwal, Sara Culver, Scott Ethersmith, Scott Gray, Sean Grove, Sean Metzger, Shamez Hermani, Shantanu Jain, Shengjia Zhao, Sherwin Wu, Shino Jomoto, Shirong Wu, Shuaiqi, Xia, Sonia Phene, Spencer Papay, Srinivas Narayanan, Steve Coffey, Steve Lee, Stewart Hall, Suchir Balaji, Tal Broda, Tal Stramer, Tao Xu, Tarun Gogineni, Taya Christianson, Ted Sanders, Tejal Patwardhan, Thomas Cunninghman, Thomas Degry, Thomas Dimson, Thomas Raoux, Thomas Shadwell, Tianhao Zheng, Todd Underwood, Todor Markov, Toki Sherbakov, Tom Rubin, Tom Stasi, Tomer Kaftan, Tristan Heywood, Troy Peterson, Tyce Walters, Tyna Eloundou, Valerie Qi, Veit Moeller, Vinnie Monaco, Vishal Kuo, Vlad Fomenko, Wayne Chang, Weiyi Zheng, Wenda Zhou, Wesam Manassra, Will Sheu, Wojciech Zaremba, Yash Patil, Yilei Qian, Yongjik Kim, Youlong Cheng, Yu~Zhang, Yuchen He, Yuchen Zhang, Yujia Jin, Yunxing Dai, and Yury Malkov. 2024{\natexlab{a}}.
\newblock \href {http://arxiv.org/abs/2410.21276} {Gpt-4o system card}.

\bibitem[{OpenAI et~al.(2024{\natexlab{b}})OpenAI, Jaech, Kalai, Lerer, Richardson, El-Kishky, Low, Helyar, Madry, Beutel, Carney, Iftimie, Karpenko, Passos, Neitz, Prokofiev, Wei, Tam, Bennett, Kumar, Saraiva, Vallone, Duberstein, Kondrich, Mishchenko, Applebaum, Jiang, Nair, Zoph, Ghorbani, Rossen, Sokolowsky, Barak, McGrew, Minaiev, Hao, Baker, Houghton, McKinzie, Eastman, Lugaresi, Bassin, Hudson, Li, de~Bourcy, Voss, Shen, Zhang, Koch, Orsinger, Hesse, Fischer, Chan, Roberts, Kappler, Levy, Selsam, Dohan, Farhi, Mely, Robinson, Tsipras, Li, Oprica, Freeman, Zhang, Wong, Proehl, Cheung, Mitchell, Wallace, Ritter, Mays, Wang, Such, Raso, Leoni, Tsimpourlas, Song, von Lohmann, Sulit, Salmon, Parascandolo, Chabot, Zhao, Brockman, Leclerc, Salman, Bao, Sheng, Andrin, Bagherinezhad, Ren, Lightman, Chung, Kivlichan, O'Connell, Osband, Gilaberte, Akkaya, Kostrikov, Sutskever, Kofman, Pachocki, Lennon, Wei, Harb, Twore, Feng, Yu, Weng, Tang, Yu, Candela, Palermo, Parish, Heidecke, Hallman, Rizzo, Gordon, Uesato,
  Ward, Huizinga, Wang, Chen, Xiao, Singhal, Nguyen, Cobbe, Shi, Wood, Rimbach, Gu-Lemberg, Liu, Lu, Stone, Yu, Ahmad, Yang, Liu, Maksin, Ho, Fedus, Weng, Li, McCallum, Held, Kuhn, Kondraciuk, Kaiser, Metz, Boyd, Trebacz, Joglekar, Chen, Tintor, Meyer, Jones, Kaufer, Schwarzer, Shah, Yatbaz, Guan, Xu, Yan, Glaese, Chen, Lampe, Malek, Wang, Fradin, McClay, Pavlov, Wang, Wang, Murati, Bavarian, Rohaninejad, McAleese, Chowdhury, Chowdhury, Ryder, Tezak, Brown, Nachum, Boiko, Murk, Watkins, Chao, Ashbourne, Izmailov, Zhokhov, Dias, Arora, Lin, Lopes, Gaon, Miyara, Leike, Hwang, Garg, Brown, James, Shu, Cheu, Greene, Jain, Altman, Toizer, Toyer, Miserendino, Agarwal, Hernandez, Baker, McKinney, Yan, Zhao, Hu, Santurkar, Chaudhuri, Zhang, Fu, Papay, Lin, Balaji, Sanjeev, Sidor, Broda, Clark, Wang, Gordon, Sanders, Patwardhan, Sottiaux, Degry, Dimson, Zheng, Garipov, Stasi, Bansal, Creech, Peterson, Eloundou, Qi, Kosaraju, Monaco, Pong, Fomenko, Zheng, Zhou, McCabe, Zaremba, Dubois, Lu, Chen, Cha, Bai, He, Zhang,
  Wang, Shao, and Li}]{openai2024openaio1card}
OpenAI, Aaron Jaech, Adam Kalai, Adam Lerer, Adam Richardson, Ahmed El-Kishky, Aiden Low, Alec Helyar, Aleksander Madry, Alex Beutel, Alex Carney, Alex Iftimie, Alex Karpenko, Alex~Tachard Passos, Alexander Neitz, Alexander Prokofiev, Alexander Wei, Allison Tam, Ally Bennett, Ananya Kumar, Andre Saraiva, Andrea Vallone, Andrew Duberstein, Andrew Kondrich, Andrey Mishchenko, Andy Applebaum, Angela Jiang, Ashvin Nair, Barret Zoph, Behrooz Ghorbani, Ben Rossen, Benjamin Sokolowsky, Boaz Barak, Bob McGrew, Borys Minaiev, Botao Hao, Bowen Baker, Brandon Houghton, Brandon McKinzie, Brydon Eastman, Camillo Lugaresi, Cary Bassin, Cary Hudson, Chak~Ming Li, Charles de~Bourcy, Chelsea Voss, Chen Shen, Chong Zhang, Chris Koch, Chris Orsinger, Christopher Hesse, Claudia Fischer, Clive Chan, Dan Roberts, Daniel Kappler, Daniel Levy, Daniel Selsam, David Dohan, David Farhi, David Mely, David Robinson, Dimitris Tsipras, Doug Li, Dragos Oprica, Eben Freeman, Eddie Zhang, Edmund Wong, Elizabeth Proehl, Enoch Cheung, Eric
  Mitchell, Eric Wallace, Erik Ritter, Evan Mays, Fan Wang, Felipe~Petroski Such, Filippo Raso, Florencia Leoni, Foivos Tsimpourlas, Francis Song, Fred von Lohmann, Freddie Sulit, Geoff Salmon, Giambattista Parascandolo, Gildas Chabot, Grace Zhao, Greg Brockman, Guillaume Leclerc, Hadi Salman, Haiming Bao, Hao Sheng, Hart Andrin, Hessam Bagherinezhad, Hongyu Ren, Hunter Lightman, Hyung~Won Chung, Ian Kivlichan, Ian O'Connell, Ian Osband, Ignasi~Clavera Gilaberte, Ilge Akkaya, Ilya Kostrikov, Ilya Sutskever, Irina Kofman, Jakub Pachocki, James Lennon, Jason Wei, Jean Harb, Jerry Twore, Jiacheng Feng, Jiahui Yu, Jiayi Weng, Jie Tang, Jieqi Yu, Joaquin~Quiñonero Candela, Joe Palermo, Joel Parish, Johannes Heidecke, John Hallman, John Rizzo, Jonathan Gordon, Jonathan Uesato, Jonathan Ward, Joost Huizinga, Julie Wang, Kai Chen, Kai Xiao, Karan Singhal, Karina Nguyen, Karl Cobbe, Katy Shi, Kayla Wood, Kendra Rimbach, Keren Gu-Lemberg, Kevin Liu, Kevin Lu, Kevin Stone, Kevin Yu, Lama Ahmad, Lauren Yang, Leo Liu,
  Leon Maksin, Leyton Ho, Liam Fedus, Lilian Weng, Linden Li, Lindsay McCallum, Lindsey Held, Lorenz Kuhn, Lukas Kondraciuk, Lukasz Kaiser, Luke Metz, Madelaine Boyd, Maja Trebacz, Manas Joglekar, Mark Chen, Marko Tintor, Mason Meyer, Matt Jones, Matt Kaufer, Max Schwarzer, Meghan Shah, Mehmet Yatbaz, Melody~Y. Guan, Mengyuan Xu, Mengyuan Yan, Mia Glaese, Mianna Chen, Michael Lampe, Michael Malek, Michele Wang, Michelle Fradin, Mike McClay, Mikhail Pavlov, Miles Wang, Mingxuan Wang, Mira Murati, Mo~Bavarian, Mostafa Rohaninejad, Nat McAleese, Neil Chowdhury, Neil Chowdhury, Nick Ryder, Nikolas Tezak, Noam Brown, Ofir Nachum, Oleg Boiko, Oleg Murk, Olivia Watkins, Patrick Chao, Paul Ashbourne, Pavel Izmailov, Peter Zhokhov, Rachel Dias, Rahul Arora, Randall Lin, Rapha~Gontijo Lopes, Raz Gaon, Reah Miyara, Reimar Leike, Renny Hwang, Rhythm Garg, Robin Brown, Roshan James, Rui Shu, Ryan Cheu, Ryan Greene, Saachi Jain, Sam Altman, Sam Toizer, Sam Toyer, Samuel Miserendino, Sandhini Agarwal, Santiago Hernandez,
  Sasha Baker, Scott McKinney, Scottie Yan, Shengjia Zhao, Shengli Hu, Shibani Santurkar, Shraman~Ray Chaudhuri, Shuyuan Zhang, Siyuan Fu, Spencer Papay, Steph Lin, Suchir Balaji, Suvansh Sanjeev, Szymon Sidor, Tal Broda, Aidan Clark, Tao Wang, Taylor Gordon, Ted Sanders, Tejal Patwardhan, Thibault Sottiaux, Thomas Degry, Thomas Dimson, Tianhao Zheng, Timur Garipov, Tom Stasi, Trapit Bansal, Trevor Creech, Troy Peterson, Tyna Eloundou, Valerie Qi, Vineet Kosaraju, Vinnie Monaco, Vitchyr Pong, Vlad Fomenko, Weiyi Zheng, Wenda Zhou, Wes McCabe, Wojciech Zaremba, Yann Dubois, Yinghai Lu, Yining Chen, Young Cha, Yu~Bai, Yuchen He, Yuchen Zhang, Yunyun Wang, Zheng Shao, and Zhuohan Li. 2024{\natexlab{b}}.
\newblock \href {http://arxiv.org/abs/2412.16720} {Openai o1 system card}.

\bibitem[{Pan et~al.(2023)Pan, Albalak, Wang, and Wang}]{pan2023logiclmempoweringlargelanguage}
Liangming Pan, Alon Albalak, Xinyi Wang, and William Wang. 2023.
\newblock \href {https://doi.org/10.18653/v1/2023.findings-emnlp.248} {Logic-{LM}: Empowering large language models with symbolic solvers for faithful logical reasoning}.
\newblock In \emph{Findings of the Association for Computational Linguistics: EMNLP 2023}, pages 3806--3824, Singapore. Association for Computational Linguistics.

\bibitem[{Purdy(1991)}]{b29}
William~C Purdy. 1991.
\newblock A logic for natural language.
\newblock \emph{Notre Dame Journal of Formal Logic}, 32(3):409--425.

\bibitem[{Singh et~al.(2020)Singh, Aggarwal, and Krishnamurthy}]{b34}
Hrituraj Singh, Milan Aggarwal, and Balaji Krishnamurthy. 2020.
\newblock \href {https://api.semanticscholar.org/CorpusID:211132947} {Exploring neural models for parsing natural language into first-order logic}.
\newblock \emph{ArXiv}, abs/2002.06544.

\bibitem[{Sourati et~al.(2022)Sourati, Venkatesh, Deshpande, Rawlani, Ilievski, Sandlin, and Mermoud}]{b12}
Zhivar Sourati, Vishnu Priya~Prasanna Venkatesh, Darshan Deshpande, Himanshu Rawlani, Filip Ilievski, H{\^o}ng-{\^A}n Sandlin, and Alain Mermoud. 2022.
\newblock \href {https://api.semanticscholar.org/CorpusID:254686058} {Robust and explainable identification of logical fallacies in natural language arguments}.
\newblock \emph{Knowledge Based Systems}, 266:110418.

\bibitem[{Stab and Gurevych(2017)}]{b2}
Christian Stab and Iryna Gurevych. 2017.
\newblock \href {https://aclanthology.org/E17-1092/} {Recognizing insufficiently supported arguments in argumentative essays}.
\newblock In \emph{Proceedings of the 15th Conference of the {E}uropean Chapter of the Association for Computational Linguistics: Volume 1, Long Papers}, pages 980--990, Valencia, Spain. Association for Computational Linguistics.

\bibitem[{Touvron et~al.(2023)Touvron, Martin, Stone, Albert, Almahairi, Babaei, Bashlykov, Batra, Bhargava, Bhosale, Bikel, Blecher, Ferrer, Chen, Cucurull, Esiobu, Fernandes, Fu, Fu, Fuller, Gao, Goswami, Goyal, Hartshorn, Hosseini, Hou, Inan, Kardas, Kerkez, Khabsa, Kloumann, Korenev, Koura, Lachaux, Lavril, Lee, Liskovich, Lu, Mao, Martinet, Mihaylov, Mishra, Molybog, Nie, Poulton, Reizenstein, Rungta, Saladi, Schelten, Silva, Smith, Subramanian, Tan, Tang, Taylor, Williams, Kuan, Xu, Yan, Zarov, Zhang, Fan, Kambadur, Narang, Rodriguez, Stojnic, Edunov, and Scialom}]{b26}
Hugo Touvron, Louis Martin, Kevin Stone, Peter Albert, Amjad Almahairi, Yasmine Babaei, Nikolay Bashlykov, Soumya Batra, Prajjwal Bhargava, Shruti Bhosale, Dan Bikel, Lukas Blecher, Cristian~Canton Ferrer, Moya Chen, Guillem Cucurull, David Esiobu, Jude Fernandes, Jeremy Fu, Wenyin Fu, Brian Fuller, Cynthia Gao, Vedanuj Goswami, Naman Goyal, Anthony Hartshorn, Saghar Hosseini, Rui Hou, Hakan Inan, Marcin Kardas, Viktor Kerkez, Madian Khabsa, Isabel Kloumann, Artem Korenev, Punit~Singh Koura, Marie-Anne Lachaux, Thibaut Lavril, Jenya Lee, Diana Liskovich, Yinghai Lu, Yuning Mao, Xavier Martinet, Todor Mihaylov, Pushkar Mishra, Igor Molybog, Yixin Nie, Andrew Poulton, Jeremy Reizenstein, Rashi Rungta, Kalyan Saladi, Alan Schelten, Ruan Silva, Eric~Michael Smith, Ranjan Subramanian, Xiaoqing~Ellen Tan, Binh Tang, Ross Taylor, Adina Williams, Jian~Xiang Kuan, Puxin Xu, Zheng Yan, Iliyan Zarov, Yuchen Zhang, Angela Fan, Melanie Kambadur, Sharan Narang, Aurelien Rodriguez, Robert Stojnic, Sergey Edunov, and Thomas
  Scialom. 2023.
\newblock \href {http://arxiv.org/abs/2307.09288} {Llama 2: Open foundation and fine-tuned chat models}.

\bibitem[{Wei et~al.(2022)Wei, Wang, Schuurmans et~al.}]{wei2022chain}
Jason Wei, Xuezhi Wang, Dale Schuurmans, et~al. 2022.
\newblock \href {http://arxiv.org/abs/2201.11903} {Chain of thought prompting elicits reasoning in large language models}.
\newblock \emph{Advances in Neural Information Processing Systems}.

\bibitem[{Williams et~al.(2018)Williams, Nangia, and Bowman}]{williams-etal-2018-broad}
Adina Williams, Nikita Nangia, and Samuel Bowman. 2018.
\newblock \href {https://doi.org/10.18653/v1/N18-1101} {A broad-coverage challenge corpus for sentence understanding through inference}.
\newblock In \emph{Proceedings of the 2018 Conference of the North {A}merican Chapter of the Association for Computational Linguistics: Human Language Technologies, Volume 1 (Long Papers)}, pages 1112--1122, New Orleans, Louisiana. Association for Computational Linguistics.

\bibitem[{Yang et~al.(2024)Yang, Xiong, Payani, Shareghi, and Fekri}]{b22}
Yuan Yang, Siheng Xiong, Ali Payani, Ehsan Shareghi, and Faramarz Fekri. 2024.
\newblock \href {https://doi.org/10.18653/v1/2024.acl-long.375} {Harnessing the power of large language models for natural language to first-order logic translation}.
\newblock In \emph{Proceedings of the 62nd Annual Meeting of the Association for Computational Linguistics (Volume 1: Long Papers)}, pages 6942--6959, Bangkok, Thailand. Association for Computational Linguistics.

\end{thebibliography}

\clearpage

\appendix
\section*{Appendix}
\section{Algorithms}


\begin{algorithm}[ht]
\small
\caption{Compiling Logical Formula to SMT}
\label{alg:foltosmt}
\KwIn{Logical formula $\mathcal{L}$ in natural language or First-Order Logic (FOL)}
\KwOut{SMT file $\mathcal{S}$ formatted for formal solvers}

\SetKwFunction{Tokenize}{Tokenize}
\SetKwFunction{ProcessTokens}{ProcessTokens}
\SetKwFunction{InfixToPrefix}{InfixToPrefix}
\SetKwFunction{UnifySort}{UnifySort}
\SetKwFunction{GenerateSMT}{GenerateSMT}

\BlankLine\BlankLine
\textbf{Step 1: Tokenize Formula}\\
$\mathcal{T} \gets$ \Tokenize{$\mathcal{L}$}  \tcp{Split $\mathcal{L}$ into tokens based on operators, parentheses, and commas}

\BlankLine
\textbf{Step 2: Process Tokens}\\
$\mathcal{P} \gets \emptyset$  \tcp{Initialize processed tokens set}

\ForEach{token $t \in \mathcal{T}$}{
    \uIf{$t$ is a predicate}{
        Identify arguments of $t$\\
        Recursively \ProcessTokens{} for arguments\\
    }
    \uElseIf{$t$ is an operator or variable}{
        Add $t$ to $\mathcal{P}$\\
    }
}

\BlankLine
\textbf{Step 3: Convert Formula to Prefix Notation} \\
$\mathcal{F}_{\text{prefix}} \gets$ \InfixToPrefix{$\mathcal{P}$}
\tcp{Transform logical formula from infix to prefix notation}

Recursively apply \InfixToPrefix{} for predicate arguments\\

\BlankLine
\textbf{Step 4: Determine Sorts} \\
$\mathcal{S}_{\text{sorts}} \gets$ \UnifySort{$\mathcal{F}_{\text{prefix}}$}  \tcp{Assign sorts for variables and predicates}

\BlankLine
\textbf{Step 5: Format Formula for SMT} \\
$\mathcal{F}_{\text{SMT}} \gets$ Parenthesize $\mathcal{F}_{\text{prefix}}$ according to SMT-LIB syntax\\

\BlankLine
\textbf{Step 6: Generate SMT File} \\
$\mathcal{S} \gets$ \GenerateSMT{$\mathcal{S}_{\text{sorts}}, \mathcal{F}_{\text{SMT}}$}\\

Include \\
\begin{itemize}
    \item \texttt{(declare-sort)} statements for sorts. \\
    \item \texttt{(declare-fun)} statements for variables and predicates. \\
    \item Negation of $\mathcal{F}_{\text{SMT}}$. \\
    \item \texttt{(check-sat)} and \texttt{(get-model)} commands. \\
\end{itemize}
\BlankLine
\Return $\mathcal{S}$  \tcp{Return the SMT file for use in formal solvers}
\end{algorithm}

\begin{algorithm}[ht]
\small
\caption{UnifySort for Predicate $A(x, y)$}
\label{alg:UNIFY}
\KwIn{Predicate $A(x, y)$ with arguments and potential instances}
\KwOut{Unified sort for predicate $A$ or an error if sorts are incompatible}

\BlankLine\BlankLine
\textbf{Step 1: Declare the Current Sort} 
Initialize the current sort of $A$: $(\text{NULL}, \text{NULL}, \text{Bool})$ 

\BlankLine
\textbf{Step 2: Process Each Instance of Predicate $A$} 
\ForEach{instance of predicate $A$}{
    
    \BlankLine
    \textbf{Step 2.1: Determine Instance Sorts} 
    \ForEach{argument $x_i$ in the instance}{
        \uIf{$x_i$ is a formula}{
            Set $\text{sort}(x_i) = \text{Bool}$ 
        }
        \uElseIf{$x_i$ is a variable}{
            Set $\text{sort}(x_i) = \text{sort}(\text{variable})$ \tcp{May be NULL} 
        }
    }
    
    \BlankLine
    \textbf{Step 2.2: Unify Current Sort with Instance Sort} 
    \ForEach{statement sort in current and instance sorts}{
        \uIf{sorts are not NULL and different}{
            Raise an error: \texttt{Incompatible sorts} 
        }
        \uElseIf{current sort is NULL and instance sort is not NULL}{
            Update current sort: $\text{current\_sort} \gets \text{instance\_sort}$ 
        }
        \uElseIf{instance sort is NULL and current sort is not NULL}{
            Update variable sort to match current sort
        }
    }
}
\end{algorithm}

\section{Prompt Examples}

Note: Additional in-context examples were removed for brevity and denoted `[...]' in the following prompts. 

\subsection{End-to-end LLM Prompts}
\begin{tcolorbox}[breakable, title=Prompt \hypertarget{ex:p1}{\examplepOne}\label{ex:p1}. Classifying with in-context examples (Few-shot)]
Logical fallacies are common errors in reasoning that undermine the logic of an argument.\\

A sentence is logically valid if and only if it is not possible for it to be false.\\

Here are some examples of classifying sentences as logical fallacies or valid sentences:\\

Example 1:\\

Input: "I met a tall man who loved to eat cheese, now I believe all tall people like cheese"

Answer: Logical Fallacy\\

[...]\\

Now, classify the following sentence. Answer with either "Logical Fallacy" or "Valid" at the start of your answer.\\

Input:
\end{tcolorbox}

\subsection{Intermediate NL2FOL Prompts}
\begin{tcolorbox}[breakable,title=Prompt \hypertarget{ex:p2}{\examplepTwo}\label{ex:p2}. Extracting claim and implication]
Here are some examples of extracting claims and implications from an input paragraph. There can be multiple claims but only one implication.\\

Input: "I met a tall man who loved to eat cheese, now I believe all tall people like cheese."
 
Output:

Claim: "A tall man loves cheese."

Implication: "All tall people like cheese."\\
 
[...]\\

Do not use any subordinating conjunctions in the implication. Replace pronouns with the appropriate nouns so that there are no pronouns. Now extract the claim and implication for the following input.\\
 
Input:
\end{tcolorbox}

\begin{tcolorbox}[breakable,title=Prompt \hypertarget{ex:p3}{\examplepThree}\label{ex:p3}. Getting referring expressions]
You are given a sentence. Referring expressions are noun phrases, pronouns, and proper names that refer to some individual objects that have some properties associated with them.
Here are some examples of finding referring expressions in a sentence:\\

Input: "A tall man loved cheese"

Referring expressions: A tall man\\

[...]\\

Now, find the referring expressions for the following input: 
\end{tcolorbox}

\begin{tcolorbox}[breakable,title=Prompt \hypertarget{ex:p4}{\examplepFour}\label{ex:p4}. Getting entity relations]
Please determine the relationship between the two entities provided below. Choose the number corresponding to the statement that best describes their relationship:\\

1. "[Entity A]" is equal to "[Entity B]".

2. "[Entity A]" is a subset of "[Entity B]".

3. "[Entity B]" is a subset of "[Entity A]".

4. "[Entity A]" is not related to "[Entity B]".\\

Instructions:

- Equality check: If the two entities are equal (case-insensitive after stripping whitespace), select statement 1.

- Subset determination: If they are not equal, assess whether one entity is a subset of the other based on general knowledge and logical reasoning.

    \hspace{2em}- If "[Entity A]" is a subset of "[Entity 
    
    \hspace{2em}B]", select statement 2.
    
    \hspace{2em}- If "[Entity B]" is a subset of "[Entity 
    
    \hspace{2em}A]", select statement 3.
    
- Unrelated entities: If none of the above statements accurately describes the relationship.\\

Here are some examples:\\

Example 1:\\

Entity A: "dogs"
    
Entity B: "animals"
    
Analysis: All dogs are animals, so "dogs" is a subset of "animals".
    
Answer: 2\\

[...]\\
    
Entities:

- Entity A: {}

- Entity B: {}\\

Your Task:

- Analyze the relationship between "Entity A" and "Entity B" based on the instructions.

- Provide only the number (1, 2, 3, or 4) that corresponds to the statement you have selected.
\end{tcolorbox}

\begin{tcolorbox}[breakable,title=Prompt \hypertarget{ex:p5}{\examplepFive}\label{ex:p5}. Getting properties (claim)]
Given a sentence, and the referring expressions of that sentence. Properties are anything that describes a relationship between two referring expressions, or they may describe a trait of a referring expression. These properties are essentially predicates in first-order logic.\\

Here are some examples of finding properties in a sentence:\\

Example 1:\\

Input sentence: A tall man loves cheese

Referring expressions: tall man: a, cheese: b

Properties: IsTall(x), LovesCheese(x)\\

[...]\\

Now extract the properties for the following input:
\end{tcolorbox}

\begin{tcolorbox}[breakable,title=Prompt \hypertarget{ex:p6}{\examplepSix}\label{ex:p6}. Getting property relations]
You are given two logical clauses. Your task is to identify whether or not the first clause entails the second clause, taking into account external knowledge or 'common sense'. Also, take into account the context from the input sentence.\\

Here are some examples:\\

Example 1:\\

Input sentence: A boy is jumping on skateboard in the middle of a red bridge. Thus, the boy does a skateboarding trick.

Clause 1: JumpsOn(boy,skateboard)

Clause 2: Does(boy, skateboarding\_trick)

Answer: ENTAILMENT\\

[...]\\

Now given the following clauses. identify whether the first clause entails the second clause.
\end{tcolorbox}

\begin{tcolorbox}[breakable,title=Prompt \hypertarget{ex:p7}{\examplepSeven}\label{ex:p7}. Retrieving FOL expression]
Given a sentence, the referring expressions of that sentence, and properties which are associated with the referring expressions. Use the given properties to convert the sentence into a first-order logical form.  Use -> to represent implies, $\&$ to represent and, | to represent or and ~ to represent negations.\\
 
Example 1:\\
 
Input Sentence: A tall man loves cheese 
 
Referring Expressions: A tall man: x 
 
Properties: IsTall(x), LovesCheese(x) 
 
Logical Form: IsTall(x) $\&$ LovesCheese(x)\\

[...]
\end{tcolorbox}

The complete set of prompt examples is available in our public code repository at \url{https://github.com/lovishchopra/NL2FOL}. We encourage readers to visit the repository for details and latest updates.

\end{document}